\pdfoutput=1

\documentclass[11pt]{article}

\usepackage[final]{acl}

\usepackage{times}
\usepackage{latexsym}
\usepackage{titletoc}

\usepackage[T1]{fontenc}

\usepackage[utf8]{inputenc}

\usepackage{microtype}

\usepackage{inconsolata}

\usepackage{graphicx}

\usepackage{multirow}
\usepackage{booktabs}
\usepackage{subfigure}
\usepackage{enumitem}
\usepackage{utfsym}
\usepackage{balance}
\usepackage{makecell}
\usepackage{amssymb}
\usepackage{algorithmic}
\usepackage{algorithm}
\usepackage{diagbox}
\newcommand\blfootnote[1]{%
  \begingroup
  \renewcommand\thefootnote{}\footnote{#1}%
  \addtocounter{footnote}{-1}%
  \endgroup
}

\title{Embracing Imperfection: Simulating Students with \\ Diverse Cognitive Levels Using LLM-based Agents}

\author{
 \textbf{Tao Wu$^1$},
 \textbf{Jingyuan Chen$^{1\dagger}$},
 \textbf{Wang Lin$^1$},
 \textbf{Mengze Li$^2$},
\\
 \textbf{Yumeng Zhu$^1$},
 \textbf{Ang Li$^1$},
 \textbf{Kun Kuang$^1$},
 \textbf{Fei Wu$^{1\dagger}$}
\\
 $^1$ Zhejiang University,
 $^2$ Hong Kong University of Science and Technology
\\
\href{mailto:twu22@zju.edu.cn}{\texttt{twu22@zju.edu.cn}},  \ \ \ 
\href{mailto:jingyuanchen@zju.edu.cn}{\texttt{jingyuanchen@zju.edu.cn}}
}

\begin{document}
\maketitle
\blfootnote{$^{\dagger}$ Corresponding authors.}
\begin{abstract}

Large language models (LLMs) are revolutionizing education, with LLM-based agents playing a key role in simulating student behavior. A major challenge in student simulation is modeling the diverse learning patterns of students at various cognitive levels. However, current LLMs, typically trained as ``helpful assistants'', target at generating perfect responses. As a result, they struggle to simulate students with diverse cognitive abilities, as they often produce overly advanced answers, missing the natural imperfections that characterize student learning and resulting in unrealistic simulations.
To address this issue, we propose a training-free framework for student simulation. We begin by constructing a cognitive prototype for each student using a knowledge graph, which captures their understanding of concepts from past learning records. This prototype is then mapped to new tasks to predict student performance. Next, we simulate student solutions based on these predictions and iteratively refine them using a beam search method to better replicate realistic mistakes. To validate our approach, we construct the \texttt{Student\_100} dataset, consisting of $100$ students working on Python programming and $5,000$ learning records. Experimental results show that our method consistently outperforms baseline models, achieving $100\%$ improvement in simulation accuracy and realism. Project page: \href{https://mccartney01.github.io/student_sim}{https://mccartney01.github.io/student\_sim}.

\end{abstract}

\section{Introduction}
\label{text_intro}
Artificial intelligence (AI) is transforming education by seamlessly integrating into teaching and learning~\cite{llm_edu_survey}. Large language models (LLMs) play a central role in this shift, excelling in tasks such as personalized tutoring \cite{tutor1}, curriculum design \cite{curriculum}, and adaptive assessment \cite{adaptive_misconceptions}.
A key way LLMs drive these advancements is through simulation, where models typically adopt the role of a student. Such simulations provide researchers a cost-effective approach to evaluate teaching strategies \cite{mathvc}, assess intelligent tutoring systems \cite{student_per_sim}, and enhance AI educational tools \cite{liusocraticlm}.

\begin{figure}[tbp]
\centering 
\includegraphics[width=1\linewidth]{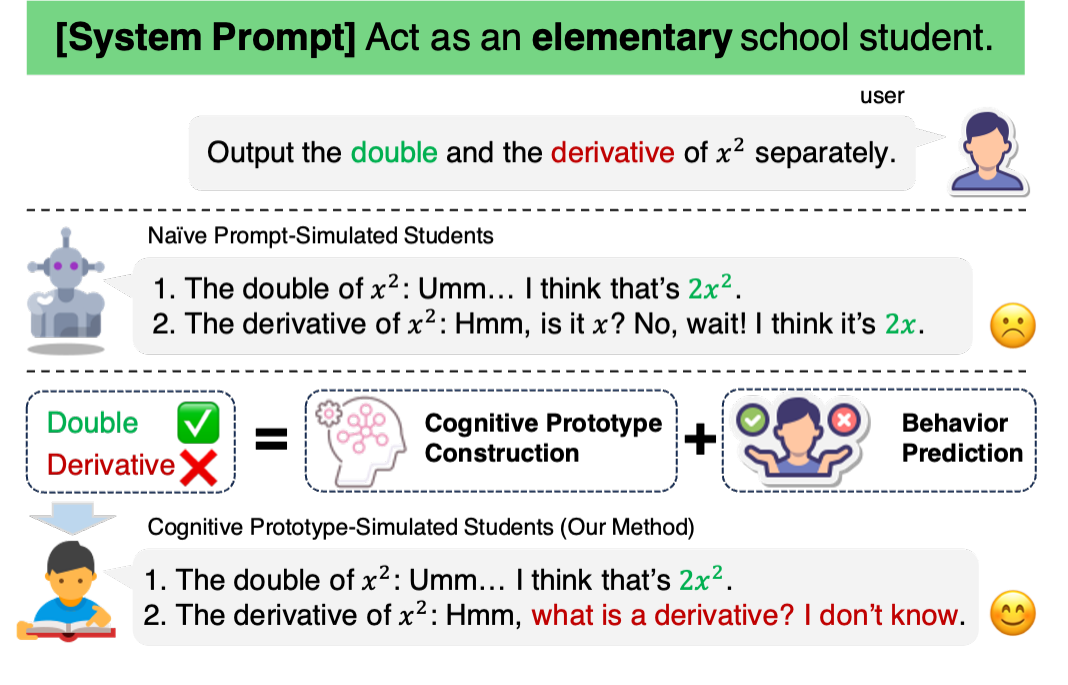}
\vspace{-0.5em}
\caption{Existing LLM-based simulations struggle to accurately replicate behaviors at varying cognitive levels and produce overly advanced responses that undermine the validity of the simulation.} 
\label{fintro1}
\vspace{-1.5em}
\end{figure}

Effective educational simulations must account for students' cognitive diversity~\cite{eduagent}. High-performing students tend to demonstrate stronger comprehension and reasoning skills, while lower-performing students may make frequent errors. Thus, for LLM-based agents to effectively simulate student behavior, they must distinguish between different cognitive levels and generate responses that accurately reflect these differences. This means not only simulating the near-``perfect'' solutions of high-achieving students but also generating ``imperfections'' of lower-performing students, including their typical mistakes.

\begin{figure}[tbp]
\centering 
\includegraphics[width=1\linewidth]{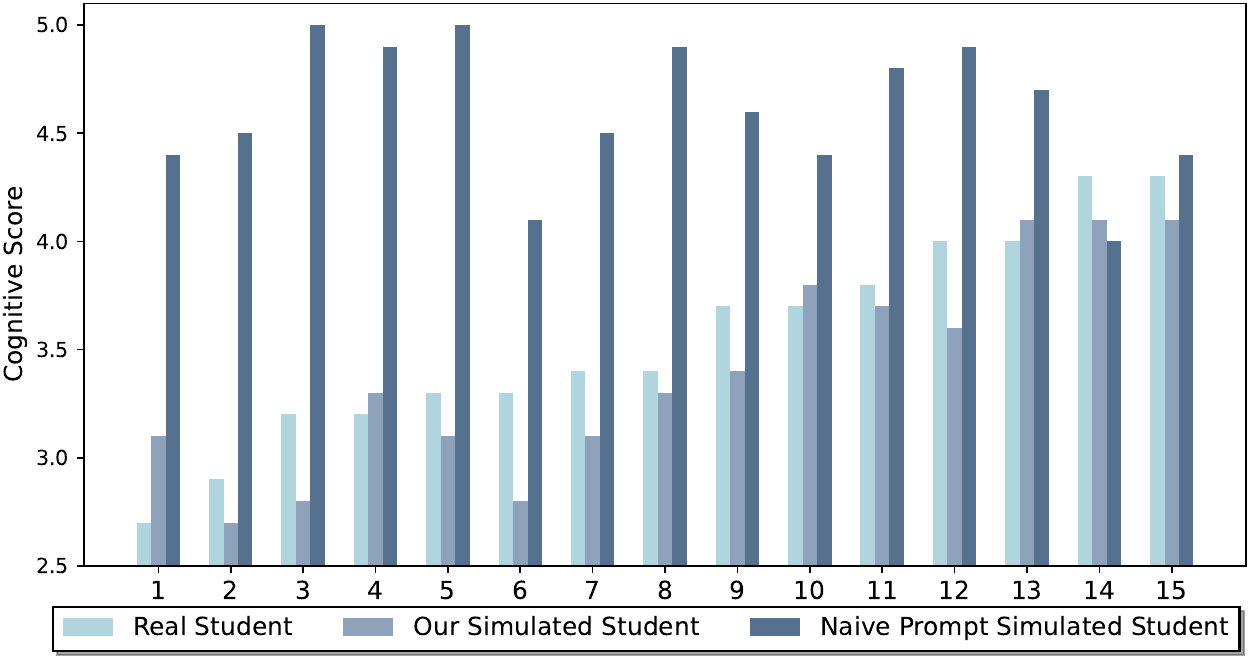}
\vspace{-0.5em}
\caption{Cognitive scores of 15 different students. The naive prompt-based simulations tend to produce overly advanced responses. In contrast, our method adaptively simulates students and produces more approximated cognitive scores.} 
\label{fintro2}
\vspace{-1em}
\end{figure}

However, recent studies highlight that current LLM-based agents struggle to replicate these ``imperfections''. They often overlook the typical mistakes made by lower-performing students and instead generate responses that are overly advanced or unrealistic~\cite{reason1, gpteach, reason3}.
As shown in Figure \ref{fintro1}, LLMs tend to overestimate the cognitive level of students with weaker abilities (\textit{e.g.}, elementary school students), failing to replicate the expected error-prone behaviors that typically occur. Figure \ref{fintro2} further illustrates this issue, showing that LLMs frequently produce responses that exceed the cognitive capabilities of these students \footnote{Details about this figure are provided in Appendix \ref{adetail_intro}.}.
This limitation likely stems from the fact that LLMs are primarily trained to provide accurate and helpful solutions \cite{threehgoal}, rather than to mimic the nuanced error patterns characteristic of student learning.

A natural approach to addressing this challenge is fine-tuning LLMs on error-rich datasets. For instance, MalAlgoPy \cite{misconceptions}  defines 20 common equation transformation errors and trains models on misconception examples to reproduce these mistakes. While such fine-tuning can encourage models to generate errors, it also risks embedding incorrect knowledge, potentially degrading overall performance. Moreover, this method overlooks the personalized nature of student learning, as mistakes should be introduced adaptively based on a student's cognitive state.

To overcome these limitations, we propose a training-free approach based on cognitive prototypes to create LLM-based agents that accurately replicate student task-solving behavior. 
Our method begins by constructing a knowledge graph-based cognitive prototype from a student's past learning records, explicitly capturing their mastery of different knowledge concepts and serving as the foundation for precise cognitive-level analysis. 
Using this prototype, we predict how students will approach new tasks in a concept-aware manner. 
Specifically, our approach maps the cognitive prototype onto the new task, enabling a deeper, knowledge-concept-level assessment of the student's mastery. This allows for more precise predictions, including whether the student can correctly solve the task and what specific errors they are likely to make.
Finally, we introduce a self-refinement approach to generate student solutions. The model continuously self-evaluates and iteratively refines its responses until they align precisely with the predicted behavior description. This not only ensures the generation of correct solutions but also improves the simulation of student mistakes, leading to more realistic, natural, and cognitively consistent student simulation.

To validate the effectiveness of our approach, we collect the \texttt{Student\_100} dataset, consisting of $5,000$ learning records from $100$ students working on Python programming tasks. Each student's records provide a unique experimental scenario for simulation. Experimental results show that our method significantly outperforms baseline models, producing more realistic and accurate student simulations.
The contributions of this paper are:
\begin{itemize}[noitemsep,nolistsep,leftmargin=*]
\item We identify the limitations of current LLMs in simulating students across different cognitive levels and collect a dataset for in-depth analysis.

\item We propose a training-free framework using cognitive prototypes to generate realistic student behaviors and solutions at various cognitive levels.

\item Extensive experimental results prove that our method overcomes the bias of behavioral simulations and can produce realistic simulations.
\end{itemize}

\section{Related Work}
\subsection{LLM-based Education Simulation}
Recent research explores using large language models (LLMs) to simulate educational roles, supporting teaching strategy evaluation~\cite{gpteach,edu_ques_gen} and AI teaching assistant development~\cite{mathvc,tutor1}. For instance, \citet{errordiag1,errordiag2} use LLMs for personalized error-correction, while \citet{student_per_sim} simulate students with different personalities in ITS. SocraticLM\cite{liusocraticlm} models students, teachers, and deans to support Socratic teaching and generate training data.
Accurately replicating student behavior, including errors, is vital. Existing methods \cite{misconceptions} improve error replication by training on error-rich data, but risk embedding incorrect knowledge. To address this, we propose a training-free pipeline that predicts and refines student behavior simulations for more realistic and natural modeling.

\subsection{Student Cognitive State Analysis}
Diagnosing students’ cognitive states and predicting their behavior are key challenges in education research. Existing methods fall into two categories: psychometrics-based and deep learning-based approaches~\cite{edujour2}. Psychometrics-based methods~\cite{psycho1,psycho2} model the relationship between performance and knowledge state using empirical response functions, while deep learning methods~\cite{deep1,deep2,dong1,dongcomprehend,lv2025debiased,zhou1,lin} leverage neural networks but rely on implicit, parameterized knowledge, making it difficult to assess mastery of specific concepts~\cite{edujour1}. This limitation restricts their effectiveness in student behavior prediction and simulation.
To address this, we propose a knowledge graph-based approach to model students’ cognitive prototypes. Our explicit, natural language-driven framework enables more precise behavior prediction and supports solution simulation.

\section{Dataset Curation}
\label{text_dataset}
Student simulation relies on the premise that a student's cognitive state remains stable over a short period. Based on this assumption, it effectively assesses the student's cognitive state using their past learning records and predicts their behavior on new tasks. The simulation then reproduces solution outcomes, such as accurately solving problems related to well-understood concepts or making reasonable errors on tasks involving less familiar concepts.

The simulation requires datasets with three key features: (1) a stable cognitive state for each student, (2) sequential task-solving records, and (3) detailed annotations, including task statements, solving behavior, and corresponding solutions. However, existing datasets often fail to meet all these criteria. Knowledge tracing datasets \cite{kt1} primarily focus on correctness, lacking textual task statements, student behavior, and solutions \cite{kt2}. In contrast, error diagnosis datasets \cite{errordiag1, errordiag2} offer rich textual information but lack annotated task-solving sequences, making it challenging to model students' cognitive states accurately.

To tackle these challenges, we develop a dataset, \texttt{Student\_100}, specifically tailored for the student simulation task. The dataset originates from an online programming platform---PTA\footnote{\href{
https://pintia.cn/}{https://pintia.cn/}}, comprising sequential programming records for each student. 
We choose programming as the task-solving scenario due to its complexity, error diversity and real-world applicability \cite{dai1,dai2}, making it a representative setting for student simulation\footnote{We provide a detailed discussion on the significance of programming tasks in Appendix \ref{app_sig_programming}.}.
To ensure a stable cognitive state, only records completed within a week are included in each sequence. Each raw record contains task statement, the solution (student-written code), and its correctness. To enhance the dataset, we recruit 10 well-trained annotators to provide detailed task descriptions and analyze student behavior (\textit{i.e.}, evaluations of the solution) for each record.

For each student sequence, we select 40 records as ``past learning records'' and 10 as ``simulation records''. The final \texttt{Student\_100} dataset comprises 100 students, each represented by 50 well-annotated task-solving records, forming a reliable foundation for student simulation tasks. Dataset statistics, a sample illustration and privacy protection are presented in Appendix \ref{adataset}.

\section{Methods}
\subsection{Overview}
Student simulation requires accurately replicating the personalized behaviors and solutions of students with varying cognitive levels during task-solving. 
To achieve this, we propose a three-stage framework for student simulation.
Firstly, given a student's $M$ past learning records $P=\{P_i\}_{1\leq i\leq M}=\{(t_i, b_i, s_i)\}_{1\leq i\leq M}$, where $t_i$, $b_i$, and $s_i$ represent tasks, behaviors, and solutions, respectively, we construct a cognitive prototype using a knowledge graph to explicitly capture the student's cognitive state (Section \ref{cognitive_prototype_construction}). Secondly, for a new task $t_j$ ($M+1\leq j\leq M+N$, where $N$ is the number of tasks to be simulated), this prototype is applied to predict the student's expected behavior $\hat{b}_j$ (Section \ref{behavior_prediction}). Finally, based on the predicted behavior, the model generates the expected solution $\hat{s}_j$ using a beam search-based self-refinement method to ensure consistency (Section \ref{solution_simulation}).

\begin{figure*}[t]
\centering 
\includegraphics[width=1\linewidth]{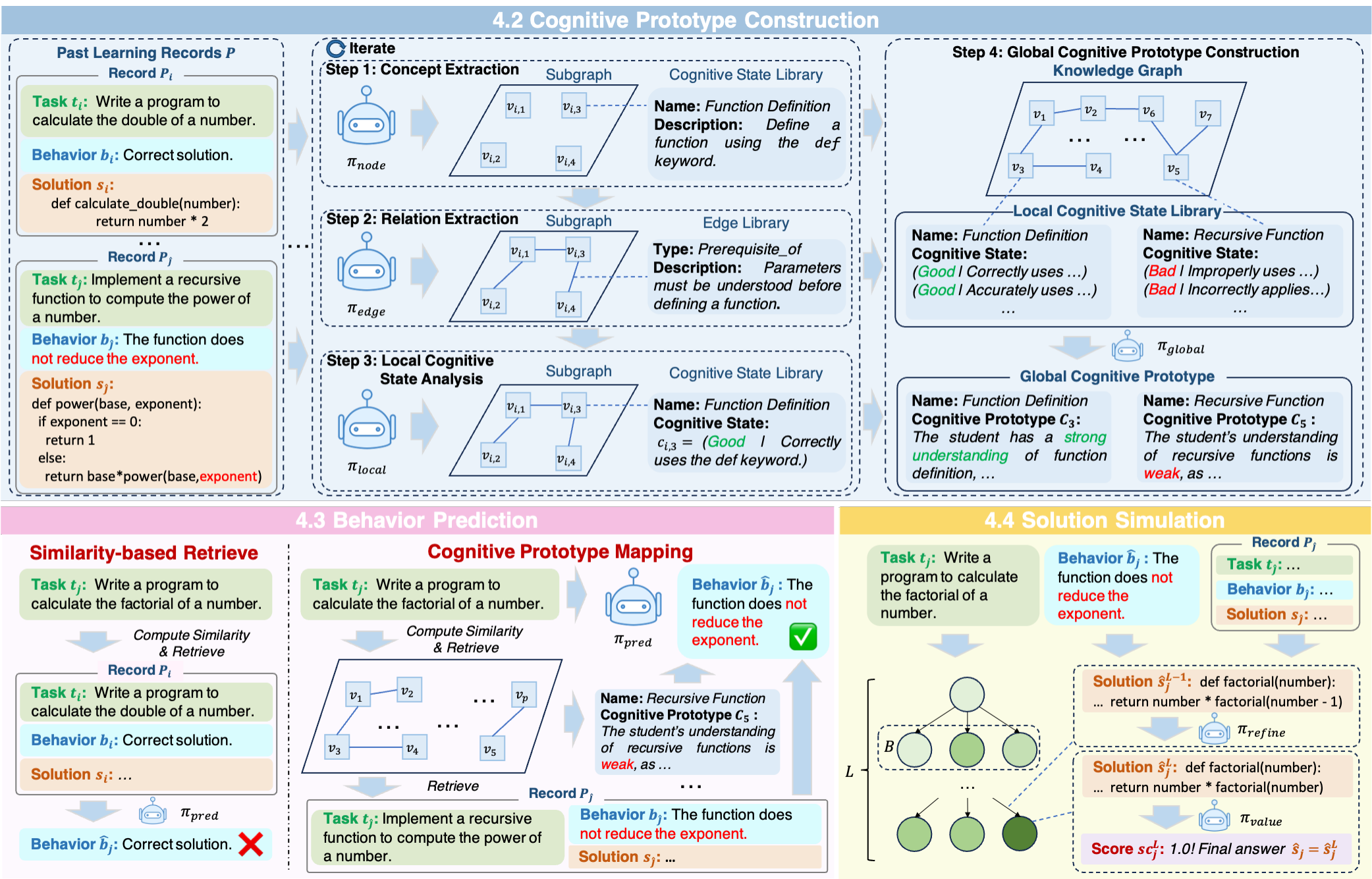
}
\caption{In the first stage, we construct a student cognitive prototype by iteratively building a knowledge graph from past learning records. This graph contains concepts, their relationships, and the local cognitive state libraries. After processing these records, we assess the student’s mastery of each concept to create a global cognitive prototype. In the second stage, we use this prototype to predict behavior for new tasks. Unlike traditional methods, which rely on superficial similarities and risk incorrect retrieval, our approach maps the cognitive prototype to the task, identifying relevant concepts for accurate predictions. In the third stage, we employ a beam search-based self-refinement process to ensure the generated solution aligns with predicted behavior, improving simulation authenticity.} 
\label{model}
\vspace{-1em}
\end{figure*}

\subsection{Cognitive Prototype Construction}
\label{cognitive_prototype_construction}
A successful student simulation requires an interpretable representation of the student’s cognitive state. Existing methods~\cite{kt2,edujour2} rely on implicit representations like neural networks, which lack interpretability and hinder behavior prediction.
To address this, we propose constructing a cognitive prototype for each student using a knowledge graph based on their past learning records. As shown in Figure \ref{model}, each record $P_i=(t_i,b_i,s_i)$ is processed iteratively to extract relevant knowledge concepts and relationships, which are organized into a natural language-based knowledge graph to form the cognitive prototype. This iterative process involves 4 key steps:

\textbf{Step 1: Concept Extraction.} 
When students complete a task, they engage with knowledge concepts at various levels, from foundational concepts (\textit{e.g.}, basic coding grammar) to advanced ones (\textit{e.g.}, algorithm design). However, relying solely on the task statement $t_i$ and student solution $s_i$ often limits the extraction to foundational concepts, as advanced ones are rarely explicitly represented.

To address this, we use the model $\pi_{desc}$ to generate a high-level task description $d_i=\pi_{desc}(t_i)$, which identifies reasoning strategies and advanced knowledge concepts required to complete the task. By integrating the task description $d_i$ with the past learning record $P_i$, we employ $\pi_{node}$ to extract comprehensive, multi-level knowledge concepts:
\begin{equation}
[v_{i,1},\ v_{i,2}, \cdots] = \pi_{node}(d_i, P_i)
\end{equation}
where $[v_{i,1},\ v_{i,2}, \cdots]$ denote the extracted concepts.

\textbf{Step 2: Relationship Extraction.} As illustrated in Figure \ref{model}, after concept extraction, we analyze their relationships to form the edges of the knowledge graph. Inspired by previous work \cite{graphusion}, we define four candidate relationships: ``\texttt{Prerequisite\_of}'', ``\texttt{Used\_for}'', ``\texttt{Hyponym\_of}'', and ``\texttt{Part\_of}''\footnote{Detailed explanation about the 4 relationships are provided in Appendix \ref{arelation}}.
The model $\pi_{edge}$ then identifies related concept pairs and classifies them into one of these types:
\begin{equation}
    [e_{i,1}, \ e_{i,2}, \cdots] = \pi_{edge}([v_{i,1},\ v_{i,2}, \cdots])
\end{equation}
where $[e_{i,1}, \ e_{i,2}, \cdots]$ denote the extracted edges.

However, overlapping edges may arise and cause conflicts, where different relationship types are assigned to the same concept pair. To resolve this, we assign an edge library to each edge and handle inconsistencies after processing all past learning records. Details are provided in Appendix \ref{aedgeconflict}. This ensures a consistent knowledge graph for further cognitive state analysis. 

\textbf{Step 3: Local Cognitive State Analysis.} The knowledge graph constructed in previous steps includes only objective concepts and their relationships, without the student's personalized cognitive state. Thus, we evaluate the student's mastery of each concept based on their task performance.
For each concept $v_{i,k}$, the model $\pi_{local}$ examines the student’s behavior and solution to detect any related mistakes, classifying mastery as either ``\texttt{Good}'' (no mistakes) or ``\texttt{Bad}’’ (mistake detected).
\begin{equation}
    [c_{i,1},\ \cdots] = \pi_{local}(d_i, P_i, [v_{i,1}, \ \cdots])
\end{equation}
where $[c_{i,1},\ \cdots]$ denote the student's local cognitive state of each concept.

As shown in Figure~\ref{model}, the local cognitive state for each concept is merged into its corresponding cognitive state library in the knowledge graph. This enriched library will support subsequent global cognitive state analysis, enabling a holistic assessment of the student's cognitive prototype.

\textbf{Step 4: Global Cognitive Prototype Construction.} 
After processing all past learning records, the knowledge graph is enriched with foundational nodes (\textit{i.e.}, knowledge concepts), edges (\textit{i.e.}, relationships), and local cognitive state analyses. To construct the global cognitive prototype, we evaluate the student’s mastery of each concept based on the cognitive state library. For each concept $v_{k}$, the model $\pi_{global}$ examines the frequency of ``\texttt{Good}'' and ``\texttt{Bad}'' classifications and generates a summary of the student’s overall cognitive prototype:
\begin{equation}
    C_k=\pi_{global}(v_{k}, [c_{k,1}, c_{k,2}, \cdots])
\end{equation}
where $C_k$ denotes the student's global cognitive state of concept $v_k$.

This approach constructs a cognitive prototype for each student based on a knowledge graph. With nodes and edges expressed in clear natural language, the prototype will effectively support subsequent behavior prediction and solution simulation.

\subsection{Behavior Prediction}
\label{behavior_prediction}
Once the student’s cognitive prototype is constructed, the next step is predicting their behavior on a new task. As shown in Figure \ref{model}, a common approach retrieves the most similar task from past learning records. However, this method often relies on textual similarity, overlooking underlying knowledge concepts. For example, when asked to ``\textit{Write a program to calculate the factorial of a number}'', the model might retrieve a similar task like ``\textit{Write a program to calculate the double of a number}'', which requires entirely different knowledge concepts and leads to inaccurate predictions.

To address this issue, we map the student's cognitive prototype onto the new task to predict their expected behavior. Specifically, for a given task $t_j$, the model $\pi_{desc}$ generates a detailed description of the concepts assessed by the task. This description is compared to the nodes in the knowledge graph to compute similarity scores, and the top-$p$ most relevant knowledge concepts are selected as the reference set, denoted as $[v_1, v_2, \cdots, v_p]$.
Using these $p$ concepts, we identify the past learning record $P_{\hat{j}}\ (1\leq \hat{j}\leq M)$ that includes the largest number of these concepts, which is deemed the most relevant task. By combining the student's overall cognitive states of these concepts with the context of $P_{\hat{j}}$, the model $\pi_{pred}$ predicts the student's expected behavior of the given task:
\begin{equation}
    \hat{b}_j=\pi_{pred}(t_j, [C_1, C_2,\cdots, C_{p}], P_{\hat{j}})
\end{equation}
where $\hat{b}_j$ denotes the predicted student behavior.

This behavior prediction leverages the student's cognitive prototype, ensuring predictions are based on deep conceptual understanding rather than superficial task statement similarities, and laying the foundation for further solution simulation.

\subsection{Solution Simulation}
\label{solution_simulation}
After predicting the expected student behavior $\hat{b}_j$ for the new task $t_j$, we combine it with the retrieved past learning record $P_{\hat{j}}$ to simulate the solution $\hat{s}_j$ that aligns with the predicted behavior. Specifically, if the predicted behavior reflects a correct approach, the solution accurately resolves the task. However, if the behavior indicates potential errors, the solution should incorporate natural, intentional mistakes consistent with the predicted behavior. As demonstrated in Figure \ref{model}, to further refine the coherence and quality of the simulated solution, we employ a beam search-based self-refinement method inspired by existing research \cite{alphamath, refine1}.
\setlength{\textfloatsep}{4pt}
\begin{algorithm}[!t]
    \caption{Beam Search-Based Self-Refinement}
    \label{alg}
    \renewcommand{\algorithmicrequire}{\textbf{Input:}}
    \renewcommand{\algorithmicensure}{\textbf{Output:}}
    \begin{algorithmic}[1]
        \REQUIRE Expected Behavior $\hat{b}_j$, Retrieved Past Learning Record $P_{\hat{j}}$, Current task $t_j$, Refine Model $\pi_{refine}$, Value Model $\pi_{value}$, Max Iteration $L$, Beam Size $B$, Threshold $\delta$  
        \ENSURE Simulated Solution $\hat{s}_j$    
        
        \STATE  $\hat{s}_j^1=\pi_{refine}(t_j, \hat{b}_j, P_{\hat{j}})$, $l=1$, $sc=0$
        \WHILE{$l\leq L$ and $sc<\delta$}
            \FOR{each $k \in [1,2,\cdots,B]$}
                \STATE $\hat{s}_j^{l,k} \sim \pi_{refine}(t_j, \hat{b}_j, P_{\hat{j}}, \hat{s}_j^{l})$
                \STATE $sc^{l,k} \sim \pi_{value}(t_j, \hat{b}_j, P_{\hat{j}}, \hat{s}_j^{l,k}))$
            \ENDFOR
            \STATE $\hat{k}=$ Argmax $(sc^{l,1}, sc^{l,2}, \cdots, sc^{l,B})$ \;
            \STATE $\hat{s}_j^{l+1}=\hat{s}_j^{l,\hat{k}}, \ sc=sc^{l,\hat{k}}, \ l\leftarrow l+1$
        \ENDWHILE
        \RETURN $\hat{s}_j= \hat{s}_j^{l}$
    \end{algorithmic}
\end{algorithm}

As shown in Algorithm \ref{alg}, the process begins with an initial weak solution $\hat{s}_j^1$, which may not fully align with the predicted behavior $\hat{b}_j$. Then the solution is iteratively refined over $L$ steps. At each iteration $l$ ($1 \leq l \leq L$), the model $\pi_{refine}$ updates the current solution $\hat{s}_j^l$ by leveraging the expected behavior and the retrieved past learning record to improve alignment. During each refinement step, $B$ candidate solutions are sampled to increase the likelihood of achieving a successful refinement.

Each candidate is evaluated by model $\pi_{value}$, which assigns a score between 0 and 1 based on its alignment with the predicted behavior. The candidate solution with highest score is selected as the output for the current iteration and becomes the starting point for the next refinement step.

The process continues until either the maximum number of iterations $L$ is reached or the value score exceeds a predefined threshold $\delta$. By employing this beam search-based self-refinement method, we generate a final simulated solution $\hat{s}_j$ that accurately aligns with the expected behavior.

\section{Experiments}
\subsection{Experiment Setup}
\label{experiment_setup}
We list the key parameter settings for our method in Table \ref{tvalue}. The same model is employed across all components of our approach (\textit{i.e.,} $\pi_{desc}$, $\pi_{node}$, $\pi_{edge}$, $\pi_{local}$, $\pi_{global}$, $\pi_{pred}$, $\pi_{refine}$, and $\pi_{value}$) to ensure consistency. Experiments are conducted using four representative LLMs—LLaMA-3.3-70B~\cite{llama33}, Claude-3.5-Sonnet~\cite{claude}, GPT-3.5~\cite{gpt35}, and GPT-4o~\cite{gpt4}—spanning different model families and scales. The temperature is fixed at 0 for reproducibility, except for $\pi_{refine}$, where diverse sampling is enabled to increase the likelihood of successful solution refinement.

\begin{table}[t]
\centering
\resizebox{0.7\linewidth}{!}{
\begin{tabular}{c|ccccccc}
\toprule
Parameter & $M$ & $N$ & $p$ & $L$ & $B$ & $\delta$ \\ \midrule
Value & 40 & 10 & 5 & 3 & 2 & 0.9 \\
 \bottomrule
\end{tabular}
}
\caption{Key parameter settings of our method.}
\label{tvalue}
\end{table}

\begin{table*}[ht]
\centering
\resizebox{\linewidth}{!}{
\begin{tabular}{c|c|ccc|ccc|ccc|ccc|ccc|ccc}
\toprule
\multicolumn{2}{c|}{\textbf{Behavior Prediction}} & \multicolumn{3}{c|}{Random} & \multicolumn{3}{c|}{Similarity} & \multicolumn{3}{c|}{Level} & \multicolumn{3}{c|}{Level+Random} & \multicolumn{3}{c|}{Level+Similarity} & \multicolumn{3}{c}{Prototype Mapping}\\
\midrule
\multicolumn{2}{c|}{\textbf{Solution Simulation}} & IO & CoT & Refine & IO & CoT & Refine & IO & CoT & Refine & IO & CoT & Refine & IO & CoT & Refine  & IO & CoT & Refine \\
\midrule\midrule

\multirow{3}[2]{*}{\textbf{\makecell[c]{LLaMA-3.3-\\70B-Instruct}}}  & \textbf{Acc} & 0.37 & 0.37 & 0.37 & 0.41 & 0.41 & 0.41 & 0.39 & 0.39 & 0.39 & 0.4 & 0.4 & 0.4 & 0.43 & 0.43 & 0.43 & \textbf{0.61} & \textbf{0.61} & \textbf{0.61}\\
& \textbf{Con\textsubscript{1}} & 2.29 & 2.29 & 2.29 & 2.45 & 2.45 & 2.45 & 2.3 & 2.3 & 2.3 & 2.23 & 2.23 & 2.23& 2.41 & 2.41 & 2.41 & \textbf{2.99} & \textbf{2.99} & \textbf{2.99}\\
& \textbf{Con\textsubscript{2}} & 1.7 & 1.79 & 1.8 & 2.07 & 2.03 & 1.99 & 2.12 & 2.2 & 1.93 & 1.75 & 1.75 & 1.87 & 2.03 & 2.03 & 2.08 & 2.52 & 2.49 & \textbf{2.69}\\
\midrule
\multirow{3}[2]{*}{\textbf{\makecell[c]{Claude-3.5-\\Sonnet}}}  & \textbf{Acc} & 0.53 & 0.53 & 0.53 & 0.61 & 0.61 & 0.61 & 0.42 & 0.42 & 0.42 & 0.49 & 0.49 & 0.49 & 0.47 & 0.47& 0.47 & \textbf{0.65} & \textbf{0.65} & \textbf{0.65} \\
& \textbf{Con\textsubscript{1}} & 2.74 & 2.74 & 2.74 & 3.03 & 3.03 & 3.03 & 2.24 & 2.24 & 2.24 & 2.51 & 2.51 & 2.51 & 2.44 & 2.44 & 2.44 & \textbf{3.09} & \textbf{3.09} & \textbf{3.09}\\
& \textbf{Con\textsubscript{2}} & 2.57 & 2.43 & 2.53 & 2.8 & 2.75 & 2.82 & 1.45 & 1.77 & 1.51 & 2.14 & 1.99 & 2.16 & 2.38 & 2.33 & 2.26 & 2.91 & 2.71 & \textbf{2.99}\\
\midrule
\multirow{3}[2]{*}{\textbf{\makecell[c]{GPT-3.5}}}  & \textbf{Acc} & 0.37 & 0.37 & 0.37 & 0.44 & 0.44 & 0.44 & 0.54 & 0.54 & 0.54 & 0.45 & 0.45 & 0.45 & 0.47 & 0.47 & 0.47 & \textbf{0.56} & \textbf{0.56} & \textbf{0.56}\\
& \textbf{Con\textsubscript{1}} & 2.36 & 2.36 & 2.36 & 2.51 & 2.51 & 2.51 & 2.88 & 2.88 & 2.88 & 2.61 & 2.61 & 2.61 & 2.66 & 2.66 & 2.66 & \textbf{2.99} & \textbf{2.99} & \textbf{2.99}\\
& \textbf{Con\textsubscript{2}} & 3.27 & 3.33 & 3.47 & 3.34 & 3.35 & 3.39 & 3.33 & 3.35 & 3.37 & 3.31 & 3.34 & 3.3 & 3.17 & 3.33 & 3.35 & 3.24 & 3.41 & \textbf{3.49}\\
\midrule
\multirow{3}[2]{*}{\textbf{\makecell[c]{GPT-4o}}}  & \textbf{Acc} & 0.45 & 0.45 & 0.45 & 0.47 & 0.47 & 0.47 & 0.44 & 0.44 & 0.44 & 0.47 & 0.47 & 0.47 & 0.47 & 0.47 & 0.47 & \textbf{0.94} & \textbf{0.94} & \textbf{0.94}\\
& \textbf{Con\textsubscript{1}} & 2.55 & 2.55 & 2.55 & 2.62 & 2.62 & 2.62 & 2.44 & 2.44 & 2.44 & 2.58 & 2.58 & 2.58 & 2.57 & 2.57 & 2.57 & \textbf{3.77} & \textbf{3.77} & \textbf{3.77}\\
& \textbf{Con\textsubscript{2}} & 2.49 & 2.45 & 2.31 & 2.55 & 2.86 & 2.65 & 2.27 & 2.5 & 2.45 & 2.27 & 2.29 & 2.11 & 2.24 & 2.36 & 2.29 & 3.32 & 3.5 & \textbf{3.65}\\
 \bottomrule
\end{tabular}
}
\caption{End-to-end comparison across 6 behavior prediction and 3 solution simulation methods. \textit{Acc} and \textit{Con\textsubscript{1}} metrics both evaluate behavior descriptions, yielding identical values for the same behavior prediction method.}
\label{tmain}

\end{table*}

\begin{table*}[ht]
\centering
\resizebox{\linewidth}{!}{
\begin{tabular}{c|ccccc|ccc|ccc|ccc|ccc}
\toprule
& \multirow{2}[2]{*}{\textbf{\makecell[c]{$\pi_{desc}$, $\pi_{node}$,\\$\pi_{edge}$, $\pi_{local}$}}} & \multirow{2}[2]{*}{\textbf{$\pi_{global}$}} & \multirow{2}[2]{*}{\textbf{$\pi_{pred}$}} & \multirow{2}[2]{*}{\textbf{$\pi_{refine}$}} & \multirow{2}[2]{*}{\textbf{$\pi_{value}$}} & \multicolumn{3}{c|}{\textbf{LLaMA-3.3-70B}} & \multicolumn{3}{c|}{\textbf{Claude-3.5-Sonnet}} & \multicolumn{3}{c|}{\textbf{GPT-3.5}} & \multicolumn{3}{c}{\textbf{GPT-4o}} \\ 
\cmidrule{7-18}
 & & &&&& \textbf{Acc} & \textbf{Con\textsubscript{1}} & \textbf{Con\textsubscript{2}} &
 \textbf{Acc} & \textbf{Con\textsubscript{1}} & \textbf{Con\textsubscript{2}}& 
 \textbf{Acc} & \textbf{Con\textsubscript{1}} & \textbf{Con\textsubscript{2}}& 
 \textbf{Acc} & \textbf{Con\textsubscript{1}} & \textbf{Con\textsubscript{2}} \\
 \midrule
 \midrule
 1 & & & &\checkmark &\checkmark & 0.41 & 2.45 & 1.99 & 0.61 & 3.03 & 2.82 & 0.44 & 2.51 & 3.39 & 0.47 & 2.62 & 2.65 \\
 2 &\checkmark & &\checkmark &\checkmark &\checkmark & 0.53 & 2.82 & 2.16 & 0.64 & 3.09 & 2.63 & 0.45 & 2.61 & 3.12 & 0.66 & 3.13 & 2.89 \\
 3 &\checkmark &\checkmark & &\checkmark &\checkmark & - & - & 1.98 & - & - & 2.52 & - & - & 3.26 & - & - & 2.7 \\
 4 &\checkmark &\checkmark &\checkmark & & & 0.61 & 2.99 & 2.49 & 0.65 & 3.09 & 2.71 & 0.56 & 2.99 & 3.41 & 0.94 & 3.77 & 3.5 \\
 5 &\checkmark &\checkmark &\checkmark &\checkmark & & 0.61 & 2.99 & 2.61 & 0.65 & 3.09 & 2.96 & 0.56 & 2.99 & 3.21 & 0.94 & 3.77 & 3.52\\
 6 &\checkmark &\checkmark &\checkmark &\checkmark &\checkmark & 0.61 & 2.99 & 2.69 & 0.65 & 3.09 & 3.09 & 0.56 & 2.99 & 3.49 & 0.94 & 3.77 & 3.65\\
 \bottomrule
\end{tabular}
}
\caption{Ablation study of each component of our method. \textit{Acc} and \textit{Con\textsubscript{1}} assess behavior prediction only and are thus unaffected by the changes in the solution simulation components (\textit{i.e.}, $\pi_{pred}$, $\pi_{refine}$, and $\pi_{value}$).}
\label{tabl}
\vspace{-1em}
\end{table*}

\noindent\textbf{Datasets.} Given the substantial computational and financial costs associated with student simulation—with around 100 different experimental settings, as discussed in Section \ref{main_end_to_end} and \ref{in-depth-analysis}, each requiring approximately 20 minutes per student and multiple API calls—we conduct our analytical experiments on a randomly selected subset of 15 students (\texttt{Student\_15}). This subset allows for comprehensive testing while ensuring the feasibility. For the main end-to-end comparison, we extend our analysis to all students in the \texttt{Student\_100}, ensuring that our results are both robust and representative as in Appendix \ref{app_student_100}.

To further validate the effectiveness of our method beyond Python programming, we additionally construct two new student groups---each comprising 5 students---for Java and C++ programming, respectively. These 10 students are built using metadata derived from the Codenet dataset \cite{codenet}, demonstrating our method's adaptability to datasets from different programming environments. Detailed experimental results are provided in Appendix \ref{app_java_cplusplus}.

\noindent\textbf{Baselines.} For behavior prediction, we compare our prototype mapping approach with five baselines: 
1) \textit{Random}, which randomly selects a record from past learning records as a reference; 
2) \textit{Similarity}, which selects a record based on task statement similarities; 
3) \textit{Level}~\cite{student_sim_emnlp}, which estimates a student's ability level based on the accuracy of their past learning records;
4) \textit{Level+Random}, which incorporates \textit{Random} and \textit{Level};
5) \textit{Level+Similarity}, which incorporates \textit{Random} and \textit{Similarity}.

For solution simulation, we compare our beam search-based self-refinement method with two baselines:
1) \textit{IO}~\cite{tot}, which requires models to directly output the solution with simulation instruct;
2) \textit{CoT}~\cite{cot}, which indicates a Chain-of-Thought approach.

\noindent\textbf{Metrics.} For behavior prediction, we first assess the model's accuracy (\textit{Acc}) in determining whether a student can correctly solve the tasks. To evaluate the alignment between generated and ground truth behavior descriptions, we introduce an LLM-based metric (\textit{Con\textsubscript{1}}) that scores consistency on a scale from 1 to 5, using evaluations from o1-mini. Similarly, solution simulation is measured with \textit{Con\textsubscript{2}}, following the same o1-mini approach as \textit{Con\textsubscript{1}}. 

Please note that since \textit{Acc} and \textit{Con\textsubscript{1}} exclusively assess the accuracy and consistency of behavior prediction, their values are invariant across different solution simulation methods in Table \ref{tmain} and \ref{tabl}. This design ensures a clean separation of evaluation for the two stages in our framework.

Please refer to Appendix \ref{aexpdetail} for experiment details and Appendix \ref{aexp} for more experimental results.

\subsection{End-to-End Performance Comparison}
\label{main_end_to_end}

With 6 behavior prediction and 3 solution simulation methods, we conduct end-to-end comparison across all 18 unique configurations. The results, shown in Table \ref{tmain}, reveal the following key insights:

1) \textbf{Superior behavior prediction}. Our prototype mapping approach significantly outperforms existing methods, emphasizing the importance of precisely capturing a student’s mastery of relevant concepts for more accurate behavior prediction.

2) \textbf{Enhanced solution simulation}. With our predicted behavior descriptions, our beam search-based self-refinement method consistently performs better, showing that iterative self-evaluation and optimization lead to more accurate solutions, consistent with the student’s cognitive ability.

Beyond these improvements, the end-to-end comparison also reveals some innovative findings:

1) \textbf{Stronger LLMs benefit more from self-refinement}. We observe that more powerful models (\textit{e.g.}, GPT-4o) exhibit greater performance gains with our method. We hypothesize that this is due to their superior self-evaluation capabilities, which allow them to generate more accurate assessments and constructive feedback throughout the refinement process. This iterative feedback mechanism systematically corrects imperfections in initial outputs, further enhancing simulation fidelity.

2) \textbf{Self-refinement relies on high-quality behavior descriptions}. While self-refinement generally improves solution simulation, we find that in some cases—particularly when behavior prediction quality is low—it underperforms compared to IO or CoT prompting. We hypothesize that this is because low-quality behavior descriptions can mislead the refinement process, causing the model to iteratively adjust solutions in the wrong direction, ultimately degrading simulation quality. This suggests that self-refinement is most effective when built on accurate behavior predictions.

\begin{figure}[tbp]
\centering 
\includegraphics[width=\linewidth]{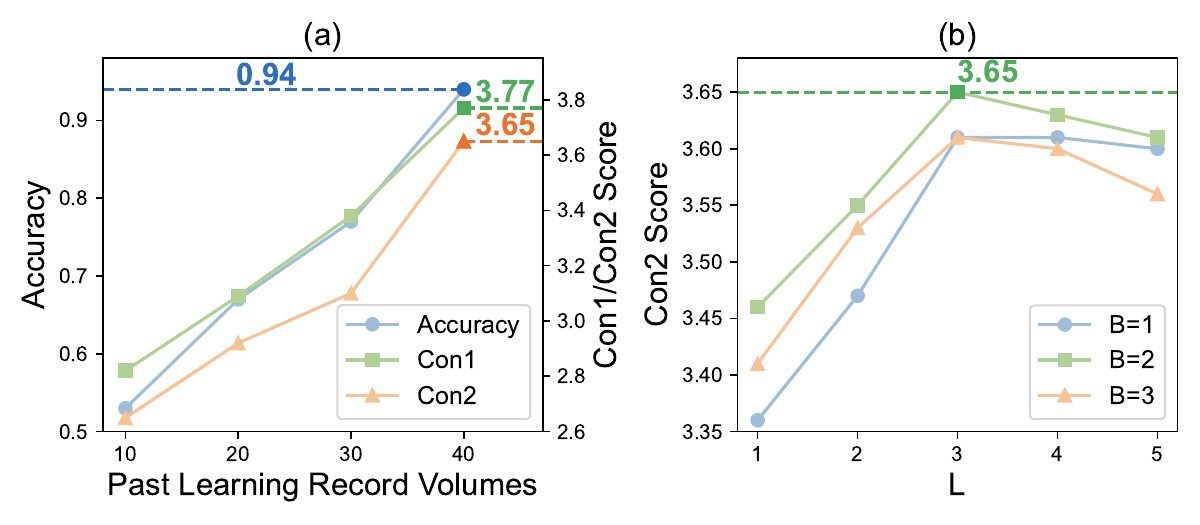}
\caption{(a) Performance on different past learning record volumes. (b) Performance on different refinement iteration $L$ and beam search sampling size $B$.} 
\label{fvolume}
\end{figure}

\begin{figure}[tbp]
\centering 
\includegraphics[width=0.6\linewidth]{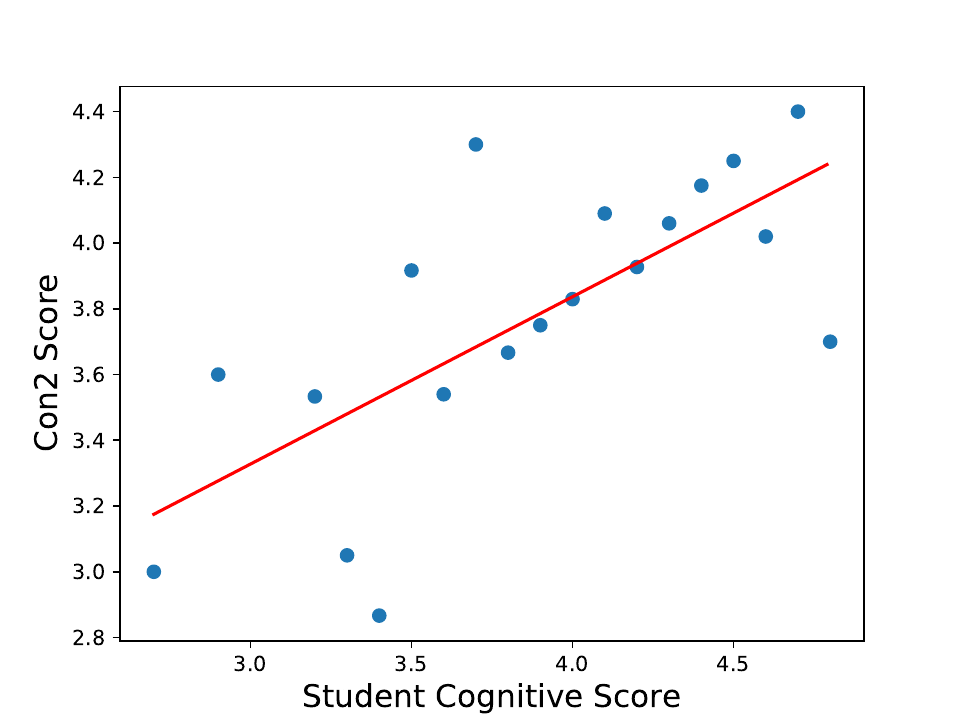}
\vspace{-0.5em}
\caption{Simulation difficulty varies across students.} 
\label{fdifferent_student}
\end{figure}

\subsection{In-Depth Analysis}
\label{in-depth-analysis}
We further validate 4 vital issues as follows.

\noindent\textbf{Effectiveness of each component.} 
We evaluate the effectiveness of each component in Table \ref{tabl}. The components $\pi_{desc}$, $\pi_{node}$, $\pi_{edge}$, and $\pi_{local}$ are considered as a whole, as ablating them individually would disrupt the entire graph-building process.
1) Removing the entire knowledge graph and relying solely on the most textually similar retrieved past learning record results in a performance drop (Row 1), validating \textit{the importance of cognitive prototype construction.}
2) Without the global cognitive prototype construction, we use only the local cognitive state library for behavior prediction, leading to a decrease in performance (Row 2). This suggests that \textit{global cognitive state construction is crucial to avoid overwhelming the model with excessive local records.}
3) Removing the behavior prediction component $\pi_{pred}$ to directly generate student solutions leads to a performance drop (Row 3), highlighting \textit{the importance of clear and accurate behavior descriptions in guiding solution simulation effectively.}
4) Removing the self-refinement process leads to poor performance in Row 4, indicating \textit{the importance of iterative refinement}. However, Row 5 shows that even with only $\pi_{refine}$, without the self-evaluation component $\pi_{value}$, performance still decreases. This underscores that \textit{self-refinement must be guided by self-evaluation to ensure the correct refinement direction.}

\noindent\textbf{Impact of past learning record volumes.} As shown in Figure \ref{fvolume} (a), increasing the number of past learning records from 10 to 40 leads to consistent improvements in both behavior prediction and solution simulation. Specifically, the behavior prediction accuracy reaches 0.94 at 40 records, marking a substantial improvement. We attribute this trend to the enhanced accuracy of the cognitive prototype, which benefits from more past learning records that captures a more comprehensive view of the student's knowledge and behavior patterns. Notably, performance continues to improve steadily at 40 records, indicating that even larger past learning records could further refine student simulation quality. Investigating the effects of larger record volumes remains a direction for future research.

\begin{table}[t]
\centering
\resizebox{\linewidth}{!}{
\begin{tabular}{c|c|c}
\toprule
\textbf{Method} & \textbf{Setting} & \textbf{Score}\\
\midrule\midrule
\multicolumn{3}{c}{\textit{LLaMA-3.3-70B-Instruct}}\\
The second best setting & Prototype Mapping + IO & 2.88 \\
Our method & Prototype Mapping + Refine & 3.08 \\
\midrule
\multicolumn{3}{c}{\textit{Claude-3.5-Sonnet}}\\
The second best setting & Prototype Mapping + IO & 3.24 \\
Our method & Prototype Mapping + Refine & 3.27 \\
\midrule
\multicolumn{3}{c}{\textit{GPT-3.5}}\\
The second best setting & Random + Refine & 3.29 \\
Our method & Prototype Mapping + Refine & 3.3 \\
\midrule
\multicolumn{3}{c}{\textit{GPT-4o}}\\
The second best setting & Prototype Mapping + CoT & 3.7 \\
Our method & Prototype Mapping + Refine & 3.75 \\
 \bottomrule
\end{tabular}
}
\vspace{-0.5em}
\caption{Human evaluation on solution simulation between our method and the second best setting.}
\label{thuman}
\end{table}

\begin{figure*}[t]
\centering 
\includegraphics[width=1\linewidth]{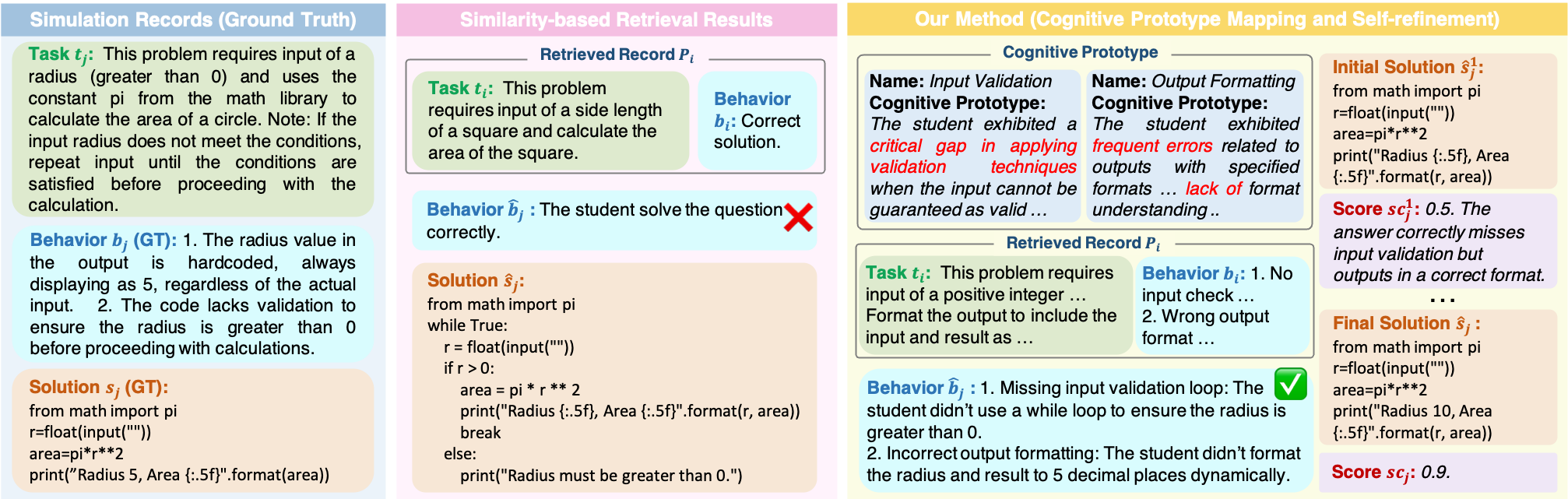}
\vspace{-1.5em}
\caption{Examples of simulated results. Similarity-based retrieval methods rely on superficial task similarities, leading to inaccurate predictions. In contrast, our method maps the cognitive prototype to relevant concepts, ensuring accuracy. The self-refinement process then iteratively adjusts the solution for precise simulation.} 
\label{fcase}
\vspace{-1em}
\end{figure*}

\noindent\textbf{Impact of self-refinement iterations and beam search sampling size.} We conduct a grid search to analyze the effects of the self-refinement iterations ($L$) and beam search sampling size ($B$), where $L$ ranges from 1 to 5 and $B$ ranges from 1 to 3. As shown in Figure \ref{fvolume} (b), the performance of solution simulation improves with an increasing iteration of refinements, highlighting the limitations of current LLMs in generating accurate student answers in a single pass. The self-refinement process effectively addresses initial weaknesses; however, when $L$ exceeds 3, performance stabilizes or slightly declines. We speculate that this is because the models tend to over-correct the solutions, which introduces deviations from realistic solutions. Therefore, we select $L = 3$.
A similar trend is observed for the beam size $B$: performance improves as $B$ increases, but the gains become negligible beyond $B = 2$. To balance performance and simulation cost, we set the sampling size to $B = 2$.

\noindent\textbf{Variability in simulation difficulty across students.} We further explore the difficulty differences in simulating individual students. We analyze the correlation between simulation quality (measured by the average \textit{Con\textsubscript{2}} score) and student cognitive ability scores (described in Appendix \ref{adataset_statistics}), as shown in Figure \ref{fdifferent_student}. The figure reveals a clear positive correlation between simulation quality and cognitive ability, indicating that simulating students with higher cognitive abilities is relatively easier.

This finding aligns with expectations---students with stronger cognitive abilities make fewer mistakes, and generating a fully correct solution is inherently easier than producing a realistic and natural simulation of a specific student's mistakes. This further highlights the challenge of accurately modeling students with lower cognitive abilities, as it requires the model to simulate not only correct responses but also plausible, individualized errors.

\subsection{Human Evaluation}
We further conduct a human evaluation study to demonstrate the effectiveness of our method. Given the substantial number of experimental settings involved in our comparisons and considering the cost of manual assessment, we compare our method against the second-best setting according to LLM evaluation results in Table \ref{tmain}.

We recruit 10 undergraduate students with solid Python programming skills to perform the evaluation. They rate each simulated solution on a 1–5 scale, consistent with the LLM-based evaluation rules, where higher scores indicate better simulation quality. The evaluation instructions provided to annotators are shown in Figure \ref{facon2}. Each solution is rated by two independent evaluators, and the final score is computed as their average.

As shown in Table \ref{thuman}, our method consistently receives higher human evaluation scores than the baseline across all model settings. This further validates the effectiveness and realism of our student simulation framework.

\subsection{Case Study}

As illustrated in Figure \ref{fcase}, similarity-based retrieval methods rely solely on textual similarities in task statements, leading to the retrieval of an inappropriate past learning record. This misleads the model into incorrectly predicting the student's behavior as ``\textit{The student solves the question correctly}'', and consequently, generating a fully correct solution, which contradicts the real student's answer.

In contrast, our method maps the constructed student cognitive prototype to the task, accurately identifying the relevant knowledge concepts the student struggles with. By retrieving a correct past learning record, the model makes an accurate behavior prediction, explicitly describing the mistakes the student is likely to make. Building on this prediction, our self-refinement process iteratively adjusts the generated solution to reflect these predicted mistakes, ultimately producing a realistic and accurate simulation of the student's solution.

\section{Conclusion}
In this paper we introduce a training-free framework for student simulation. Our approach begins by constructing a cognitive prototype for each student based on past learning records to predict their behavior. It then employs a beam search-based self-refinement process to progressively improve the quality of simulated solutions. Experiments show that our framework simulates realistic and diverse students, enhancing the realism and utility of AI in education.

\clearpage
\section*{Limitations}
In this section, we discuss the limitations of our work as follow:
\begin{itemize}
    \item While our current validation is constrained to programming domains, primarily due to the accessibility of relevant data, the underlying simulation framework is theoretically applicable to a broader range of educational subjects, such as mathematics \cite{huang2025autogeo}. We consider this a promising direction for future work.
    \item Our current simulation primarily focuses on textual and behavioral patterns without explicitly incorporating multimodal signals, such as visual or auditory cues \cite{wusemantic, multimodal2}that may also influence students' cognitive processes. While this unimodal setting enables clearer analysis of reasoning behaviors, we acknowledge that a more comprehensive understanding of student performance may benefit from integrating multimodal data. Exploring this dimension constitutes a valuable direction for future work.
\end{itemize}

\section*{Ethics Statement}
As the use of large language models (LLMs) in education grows, it is crucial to consider the ethical implications of deploying such systems to simulate human-like student behaviors. While our work explores the potential of LLMs for simulating student cognitive states and generating educational scenarios, this research remains an early-stage exploration and is conducted solely within controlled experimental settings.

The simulated students in this study are designed for research and methodological evaluation rather than real-world teaching applications. These simulations are not intended to replace human students or educators but to provide insights into model capabilities. We strongly emphasize that any deployment of such models in educational contexts should involve careful oversight by educators to prevent misuse and to ensure ethical compliance.

Future work will focus on further improving the accuracy and realism of these simulations, reducing potential biases, and incorporating diverse perspectives to enhance their applicability. We remain committed to addressing ethical challenges and ensuring that our work supports responsible and beneficial advancements in AI-driven education.

\section*{Acknowledgements}
This research was partially supported by grants from the National Natural Science Foundation of China (No.62037001, No.62307032), and the ``Pioneer'' and ``Leading Goose'' R\&D Program of Zhejiang under Grant No. 2025C02022.

\bibliography{main}

\appendix
In this appendix, we present the following content:

\startcontents
\printcontents{}{1}{}

\begin{figure}[t]
\centering 
\includegraphics[width=\linewidth]{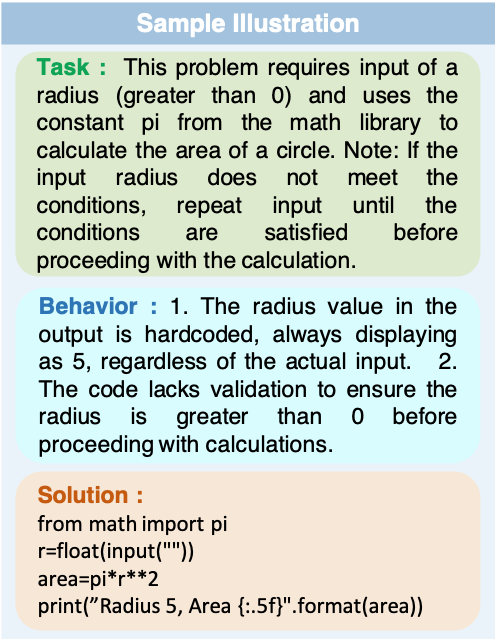}
\caption{A sample illustration in our dataset.} 
\label{fadataset_illus}
\end{figure}

\section{Dataset Statistics and Details}
\label{adataset}
\subsection{Privacy Protection}
To ensure privacy protection, we excluded all personal and sensitive data, such as student names and email addresses, retaining only anonymized unique student IDs as individual identifiers. Furthermore, we confirmed that the use of user-generated data was explicitly authorized during the registration process, as outlined in the platform's terms of service and privacy policy.

\subsection{Dataset Statistics}
\label{adataset_statistics}
In Section \ref{text_dataset}, we construct a student simulation dataset, \texttt{Student\_100}, comprising 100 students, each with 40 past learning records and 10 simulation records. Each record includes a Python programming task, the student's solving behavior (indicating correct or incorrect completion along with error analysis), and the corresponding solution (\textit{i.e.}, student-written code). A sample is illustrated in Figure \ref{fadataset_illus}.

\begin{figure}[tbp]
\centering 
\includegraphics[width=\linewidth]{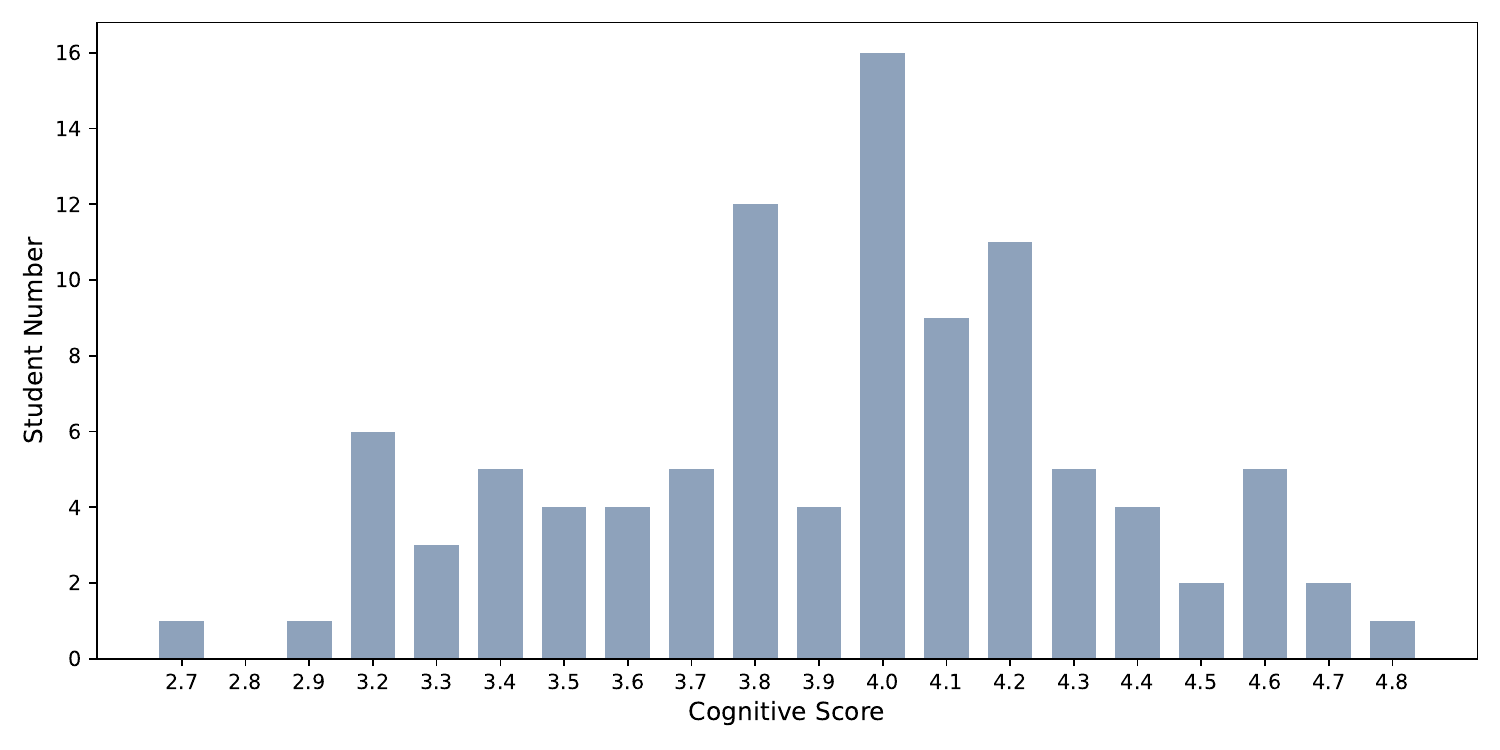}
\caption{The distribution of student cognitive scores in our dataset.} 
\label{fadataset_statics}
\end{figure}

We further analyze the distribution of these students' cognitive level.
Specifically, we use LLMs to score each student's solutions in the simulation records on a scale of 1 to 5, where higher scores indicate greater correctness. The scoring is evaluated using the o1-mini model, and the API version is o1-mini-2024-09-12. After scoring all solutions for a student, we calculate their average score to represent their cognitive ability. The distribution of these cognitive scores across the dataset is visualized in Figure \ref{fadataset_statics}.

\begin{figure}[tbp]
\centering 
\includegraphics[width=\linewidth]{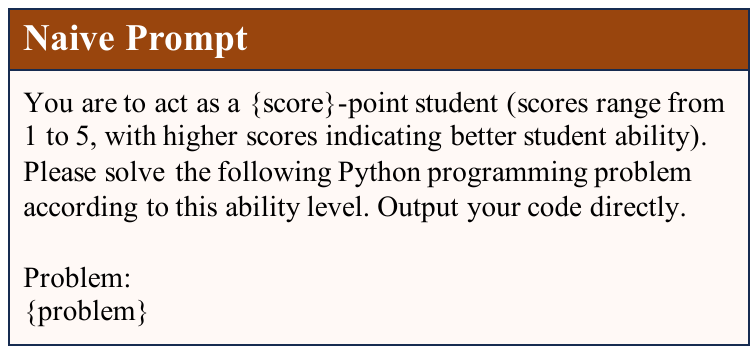}
\caption{Details of the naive prompt.} 
\label{fa_naive_prompt}
\end{figure}

\subsection{Significance of Programming Task for Student Simulation}
\label{app_sig_programming}
In this section we would like to emphasize that programming tasks are sufficiently representative for the student simulation task for several reasons:
\begin{itemize}
    \item Complexity and multi-leveled nature. Programming tasks inherently involve multiple levels of knowledge concepts, such as syntax, logical reasoning, and algorithm design. These tasks test not only a student's understanding of individual concepts but also their ability to integrate and apply these concepts to solve complex problems. This makes programming tasks an ideal scenario for modeling and simulating student cognitive states.
    \item Real-world applicability. Programming is a widely taught subject with high relevance in modern education and professional settings. Simulating student performance on programming tasks can provide practical insights into teaching strategies and learning patterns, making this work impactful in real-world educational contexts.
    \item Error diversity. Students often encounter diverse and challenging errors during programming, including syntax errors, logic errors, and runtime errors. These errors vary significantly in complexity, providing a rich testing ground for evaluating fine-grained behavior prediction and solution simulation.
    \item Interactive and iterative nature. Programming tasks typically require students to iteratively refine their solutions, reflecting a natural and realistic problem-solving process. This aligns closely with the goals of student simulation, where iterative behavior prediction and solution refinement are central.
\end{itemize}

\begin{figure}[tbp]
\centering 
\includegraphics[width=\linewidth]{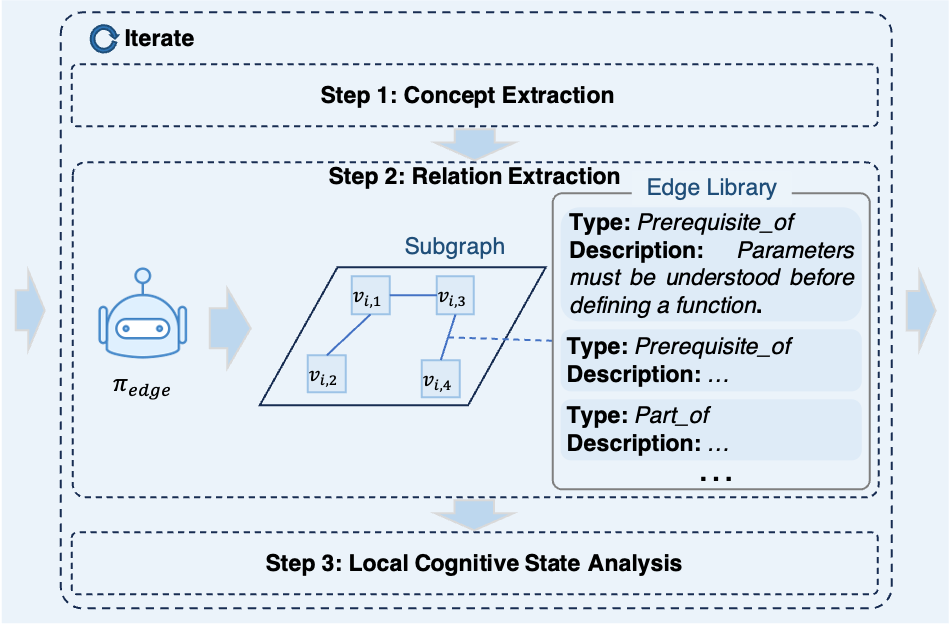}
\caption{A detailed illustration of the edge library used in the cognitive prototype construction process. During the iterative process, each identified relationship is recorded in the corresponding edge's library. After processing all learning records, the relationship type that appears most frequently in each edge's library is selected as the final type for that edge.} 
\label{faedge}
\end{figure}

We believe these characteristics make programming tasks a strong representative scenario for the student simulation problem.

\subsection{Data Generalizability}
The data used in this study was collected from an online programming platform similar to LeetCode.
Despite the platform-specific origin of the data, it has been preprocessed to ensure it is not inherently tied to any particular platform. The preprocessing includes standardizing the format of problem statements, user solutions, and associated metadata, which makes the data format highly generalizable. As a result, similar data from other platforms (\textit{e.g.}, LeetCode) can be processed into the same format, provided access to user submission records is available. This demonstrates the broader applicability of the proposed method beyond the platform used in this study.

\section{Details on Figure \ref{fintro2}}
\label{adetail_intro}
In Section \ref{text_intro}, we use Figure \ref{fintro2} to illustrate that LLMs with naive prompts consistently generate correct answers, failing to simulate students with varying abilities, particularly those with weaker performance. In contrast, our simulation method more accurately reflects the distribution of student abilities.

The experiment proceeds as follows: we use the 15 students in the random selected \texttt{Student\_15} subset (detailed in Section \ref{experiment_setup}) to conduct this experiment. As explained in Appendix \ref{adataset_statistics}, each student in the dataset is assigned a cognitive score evaluated by the LLM.
This score serves as the basis for constructing naive prompts, guiding the model to respond in alignment with the corresponding ability level. The specific prompt is shown in Figure \ref{fa_naive_prompt}.

Using this naive prompt, the model generates simulated solutions. We then apply the scoring process described in Appendix \ref{adataset_statistics} to evaluate both the solutions generated by naive prompts and those produced by our method. The average score for each simulated student is calculated as their cognitive score. The results, presented in Figure \ref{fintro2}, reveal that naive prompts consistently yield overly correct answers, failing to replicate weaker abilities. By contrast, our method closely matches the actual distribution of student abilities, providing a more realistic simulation.

\begin{figure}[tbp]
\centering 
\includegraphics[width=\linewidth]{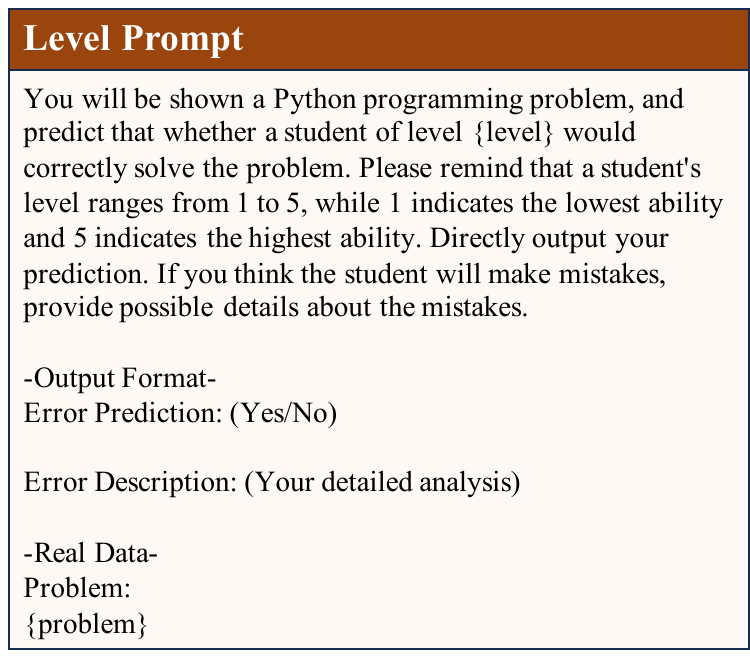}
\caption{Details of the Level prompt.} 
\label{falevel}
\end{figure}

\section{Method Details}

\subsection{Details on Concept Relationships}
\label{arelation}
In Step 2 of Section \ref{cognitive_prototype_construction}, we describe the construction of edges in the knowledge graph and define four types of relationships between the nodes (\textit{i.e.}, concepts). The following provides an overview of these relationships:

\begin{itemize}
    \item \texttt{Prerequisite\_of}: This relationship indicates that one concept is either a defining characteristic of another or a necessary prerequisite of it. For example, ``The ability to code is a prerequisite of software development.''
    \item \texttt{Used\_for}: This relationship signifies that one concept functions as a tool, resource, or method to achieve another. For instance, ``Mathematics is used for solving engineering problems.''
    \item \texttt{Hyponym\_of}: This hierarchical relationship shows that one concept is a specific instance or subtype of another. For example, ``A rectangle is a hyponym of a polygon.''
    \item \texttt{Part\_of}: This compositional relationship denotes that one concept is a component or integral part of a larger whole. For example, ``A wheel is part of a car.''
\end{itemize}

\begin{figure}[tbp]
\centering 
\includegraphics[width=\linewidth]{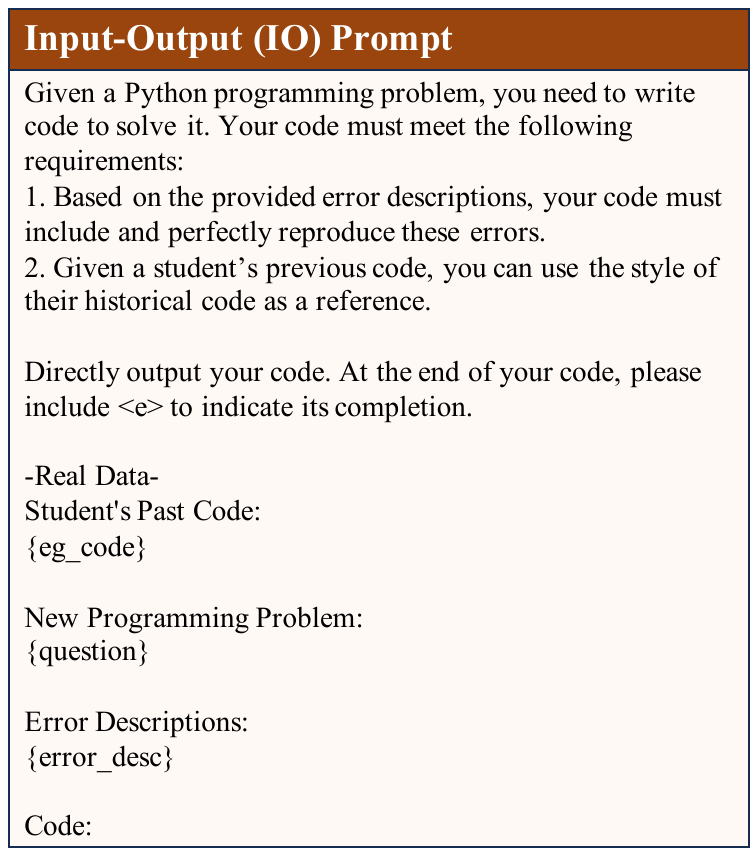}
\caption{Details of the Input-Output (IO) prompt.} 
\label{faio}
\end{figure}

\subsection{Edge Category Conflict Resolving}
\label{aedgeconflict}
In Step 2 of Section \ref{cognitive_prototype_construction}, we construct edges in the knowledge graph by defining four types of relationships between nodes (\textit{i.e.}, concepts). Since edges are extracted iteratively, overlapping edges may arise, potentially leading to conflicts where different relationship types are assigned to the same concept pair. To address this, we assign an edge library to each edge, as shown in Figure \ref{faedge}. Each identified relationship is added to the corresponding edge's library. After processing all learning records, we resolve conflicts by selecting the most frequently assigned relationship type in the edge library as the final type for each edge.

\subsection{Scalability and Computational Feasibility}
The scalability and computational feasibility of constructing the graph in our framework are key considerations. To address potential concerns about large-scale graphs, it is important to clarify that our approach is computationally manageable for the following reasons:

\begin{figure}[tbp]
\centering 
\includegraphics[width=\linewidth]{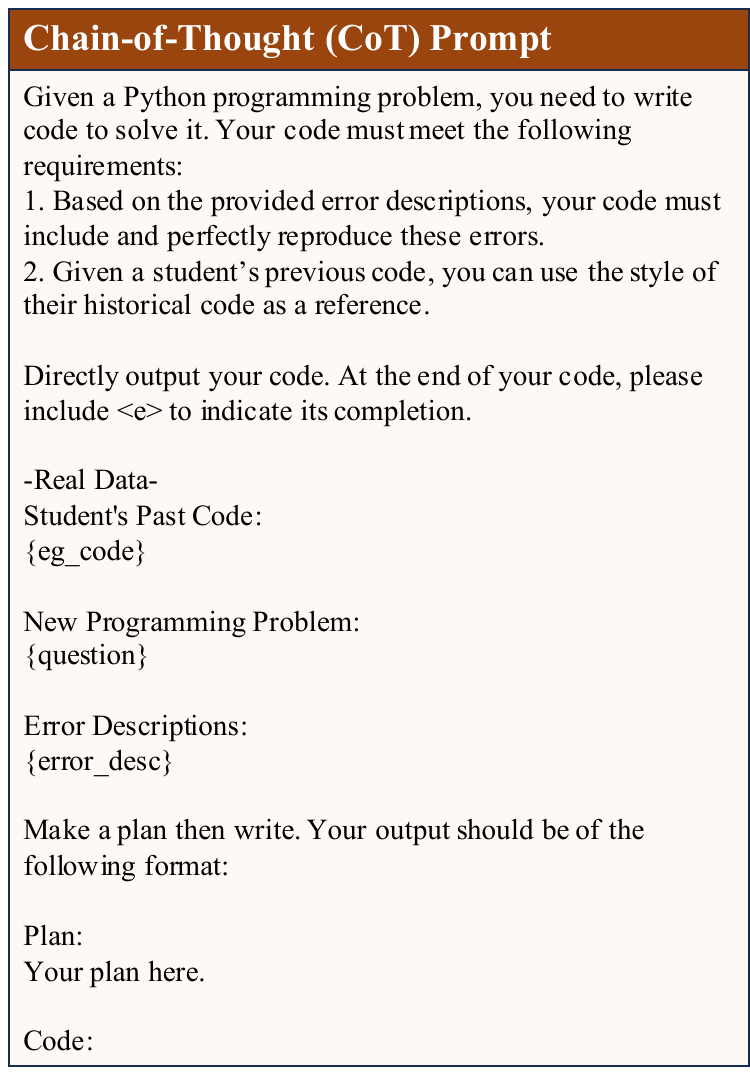}
\caption{Details of the Chain-of-Thought (CoT) prompt.} 
\label{facot}
\end{figure}

\begin{figure*}[tbp]
\centering 
\includegraphics[width=\linewidth]{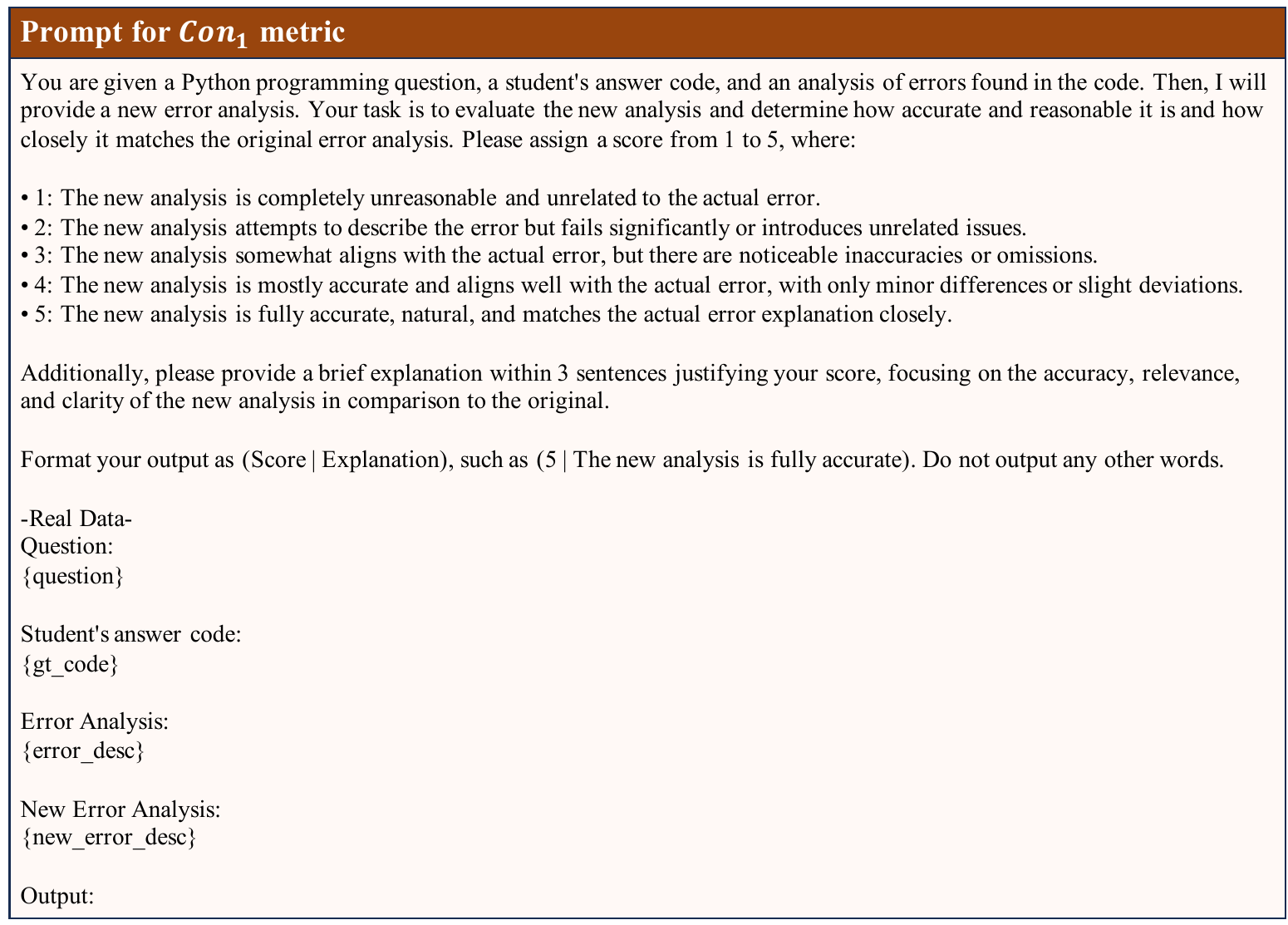}
\caption{Details of the prompt for Con\textsubscript{1} metric.} 
\label{facon1}
\end{figure*}

\textbf{Upper limit on local knowledge state analyses}. The construction of the global cognitive prototype (Step 4) evaluates a student's overall mastery of each knowledge concept by aggregating local cognitive state analyses from past learning records. The number of local analyses for each concept is directly bounded by the number of past records. In the experiments, the maximum number of local analyses is 17, which remains within a reasonable computational range. Additionally, the number of past learning records is constrained to ensure stability in the student's cognitive state, an assumption critical to the student simulation task. Specifically, only records from a one-week period are considered, during which a student typically solves no more than 50 problems. This limitation ensures that the number of local analyses remains computationally manageable.

\textbf{Upper limit on knowledge concepts}. The relationship extraction process (Step 2) focuses on analyzing the relationships between knowledge concepts within each task to form the edges of the knowledge graph. To avoid redundancy and ensure manageability, the number of extracted knowledge concepts per task is limited to 15. This restriction makes the relationship extraction process computationally efficient, well within the capabilities of large language models.

\subsection{Error Attribution and Traceability}
\label{aerror_trace}
To address the challenge of error attribution and tracing within the pipeline, our system logs each model invocation and saves the outputs of all intermediate stages. This design ensures full traceability, allowing for systematic failure analysis. When an error occurs in the final solution simulation, it is possible to trace outputs from previous stages, such as behavior prediction or retrieval, to pinpoint the root cause. This traceability supports targeted debugging and error analysis, isolating the impact of each stage on the system’s overall performance.

\begin{figure*}[t]
\centering 
\includegraphics[width=\linewidth]{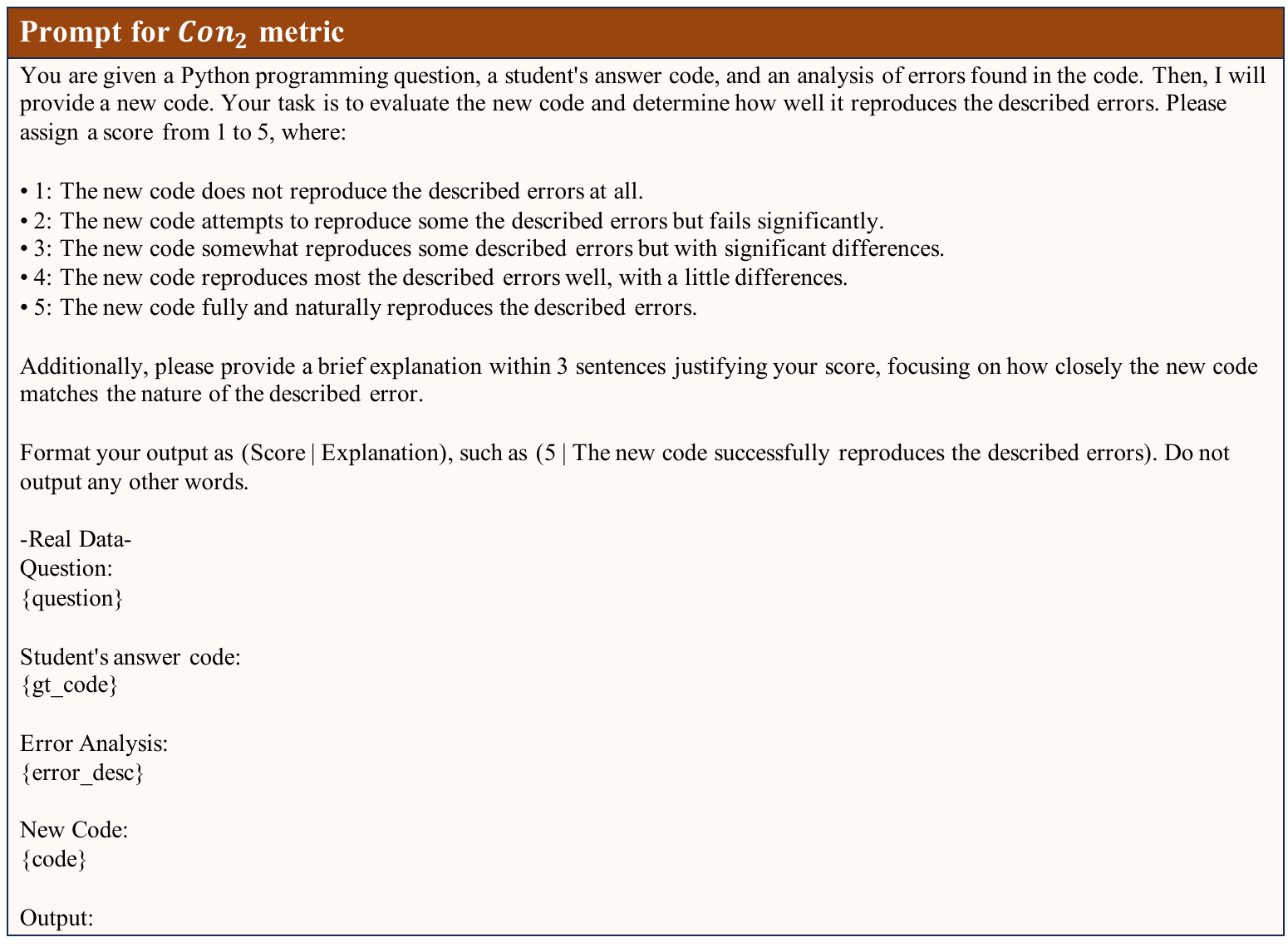}
\caption{Details of the prompt for Con\textsubscript{2} metric.} 
\label{facon2}
\end{figure*}

\section{Experiment Details}
\label{aexpdetail}
\subsection{Model Details}
\textbf{LLaMA-3.3-70B}, released by Meta AI in December 2024, is a 70-billion-parameter multilingual large language model designed to advance AI applications in both research and industry. The Llama 3.3 instruction tuned text only model is optimized for multilingual dialogue use cases and outperforms many of the available open source and closed chat models on common industry benchmarks. The test model version is LLaMA-3.3-70B-Instruct.

\noindent\textbf{Claude} is a large language model developed by Anthropic, designed to generate human-like text and assist with a wide range of tasks, including answering questions, drafting content, and engaging in conversational interactions. Built with a focus on safety and alignment, Claude leverages advanced AI techniques to ensure responsible and effective communication, making it suitable for both casual and professional use. The tested API version is claude-3-5-sonnet-20241022.

\noindent\textbf{GPT-3.5} is an advanced language model developed by OpenAI, designed to generate coherent and contextually relevant text across various domains. It builds on the capabilities of its predecessor, GPT-3, with improved understanding and responsiveness, making it highly effective for tasks like content creation, coding assistance, and conversational AI. Known for its versatility, GPT-3.5 is widely used in applications requiring natural language understanding and generation. The tested API version is gpt-3.5-turbo.

\noindent\textbf{GPT-4o} is OpenAI's generation language model, offering enhanced reasoning, creativity, and contextual understanding compared to its predecessors. It excels at complex tasks, such as advanced problem-solving, nuanced content creation, and in-depth conversational interactions. With improved alignment and multimodal capabilities, GPT-4o is designed to be more reliable, accurate, and adaptable across a wide range of applications. The tested API version is gpt-4o-2024-11-20.

\begin{table*}[ht]
\centering
\resizebox{\linewidth}{!}{
\begin{tabular}{c|c|ccc|ccc|ccc|ccc|ccc|ccc}
\toprule
\multicolumn{2}{c|}{\textbf{Behavior Prediction}} & \multicolumn{3}{c|}{Random} & \multicolumn{3}{c|}{Similarity} & \multicolumn{3}{c|}{Level} & \multicolumn{3}{c|}{Level+Random} & \multicolumn{3}{c|}{Level+Similarity} & \multicolumn{3}{c}{Prototype Mapping}\\
\midrule
\multicolumn{2}{c|}{\textbf{Solution Simulation}} & IO & CoT & Refine & IO & CoT & Refine & IO & CoT & Refine & IO & CoT & Refine & IO & CoT & Refine  & IO & CoT & Refine \\
\midrule\midrule

\multirow{3}[2]{*}{\textbf{\makecell[c]{LLaMA-3.3-\\70B-Instruct}}}  & \textbf{Acc} & 0.29 & 0.29 & 0.29 & 0.35 & 0.35 & 0.35 & 0.34 & 0.34 & 0.34 & 0.32 & 0.32 & 0.32 & 0.36 & 0.36 & 0.36 & \textbf{0.67} & \textbf{0.67} & \textbf{0.67}\\
& \textbf{Con\textsubscript{1}} & 2.26 & 2.26 & 2.26 & 2.44 & 2.44 & 2.44 & 2.28 & 2.28 & 2.28 & 2.24 & 2.24 & 2.24 & 2.43 & 2.43 & 2.43 & \textbf{3.35} & \textbf{3.35} & \textbf{3.35}\\
& \textbf{Con\textsubscript{2}} & 2.01 & 2.02 & 2.03 & 2.22 & 2.23 & 2.16 & 2.65 & 2.38 & 1.79 & 2.21 & 2.02 & 2.02 & 2.47 & 2.31 & 2.14 & 2.75 & 2.73 & \textbf{2.89}\\
\midrule

\multirow{3}[2]{*}{\textbf{\makecell[c]{Claude-3.5-\\Sonnet}}} & \textbf{Acc} &  0.49 & 0.49 & 0.49 & 0.55 & 0.55 & 0.55 & 0.27 & 0.27 & 0.27 & 0.35 & 0.35 & 0.35 & 0.36 & 0.36 & 0.36 & \textbf{0.71} & \textbf{0.71} & \textbf{0.71}\\
& \textbf{Con\textsubscript{1}} & 2.87 & 2.87 & 2.87 & 3.07 & 3.07 & 3.07 & 1.81 & 1.81 & 1.81 & 2.25 & 2.25 & 2.25 & 2.31 & 2.31 & 2.31 & \textbf{3.5} & \textbf{3.5} & \textbf{3.5}\\
& \textbf{Con\textsubscript{2}} & 2.93 & 2.77 & 2.77 & 3.1 & 3.01 & 3.01 & 1.62 & 1.72 & 1.65 & 2.41 & 2.28 & 2.27 & 2.6 & 2.56 & 2.46 & 3.22 & 3.32 & \textbf{3.38}\\
\midrule

\multirow{3}[2]{*}{\textbf{\makecell[c]{GPT-3.5}}} & \textbf{Acc} & 0.37 & 0.37 & 0.37 & 0.36 & 0.36 & 0.36 & 0.5 & 0.5 & 0.5 & 0.42 & 0.42 & 0.42 & 0.45 & 0.45 & 0.45 & \textbf{0.51} & \textbf{0.51} & \textbf{0.51}\\
& \textbf{Con\textsubscript{1}} & 2.47 & 2.47 & 2.47 & 2.47 & 2.47 & 2.47 & 2.87 & 2.87 & 2.87 & 2.66 & 2.66 & 2.66 & 2.73 & 2.73 & 2.73 & \textbf{2.96} & \textbf{2.96} & \textbf{2.96}\\
& \textbf{Con\textsubscript{2}} & 3.47 & 3.56 & 3.61 & 3.53 & 3.51 & 3.57 & 3.54 & 3.58 & 3.59 & 3.47 & 3.53 & 3.59 & 3.34 & 3.58 & 3.58 & 3.46 & 3.58 & \textbf{3.71}\\
\midrule

\multirow{3}[2]{*}{\textbf{\makecell[c]{GPT-4o}}}  & \textbf{Acc} & 0.37 & 0.37 & 0.37 & 0.42 & 0.42 & 0.42 & 0.4 & 0.4 & 0.4 & 0.55 & 0.55 & 0.55 & 0.55 & 0.55 & 0.55 & \textbf{0.89} & \textbf{0.89} & \textbf{0.89}\\
& \textbf{Con\textsubscript{1}} & 2.54 & 2.54 & 2.54 & 2.71 & 2.71 & 2.71 & 2.45 & 2.45 & 2.45 & 3.01 & 3.01 & 3.01 & 3.01 & 3.01 & 3.01 & \textbf{3.9} & \textbf{3.9} & \textbf{3.9}\\
& \textbf{Con\textsubscript{2}} & 2.73 & 2.89 & 2.57 & 2.85 & 3.08 & 2.88 & 2.42 & 2.85 & 1.83 & 2.12 & 2.65 & 2.17 & 2.48 & 2.73 & 2.27 & 3.61 & 3.65 & \textbf{3.83}\\
 \bottomrule
\end{tabular}
}
\caption{End-to-end comparison across 6 behavior prediction and 3 solution simulation methods on the whole 100 students in \texttt{Student\_100}.}
\label{tmain_app}
\end{table*}

\subsection{Baseline Details}
\label{abaseline}
For behavior prediction, we compare our prototype mapping approach with five baselines: 
\begin{itemize}
    \item \textit{Random}, which randomly selects a record from past learning records as a reference; 
    \item \textit{Similarity}, which selects a record based on task statement similarities;
    \item \textit{Level}~\cite{student_sim_emnlp}, which estimates a student's ability level based on the accuracy of their past learning records. Specifically, we first compute the accuracy of a student's past learning records and then normalize it to a range of 1 to 5, where 5 represents the highest cognitive level. As mentioned in~\citet{student_sim_emnlp}, this relative cognitive level simulation approach yields better results. We incorporate this level into the prompt to indicate the student's proficiency, enabling the model to perform behavior prediction accordingly. The prompt is shown in Figure \ref{falevel}.
    \item \textit{Level+Random}, which incorporates \textit{Random} and \textit{Level}. We add the randomly selected past learning record into the \textit{Level} prompt for behavior prediction;
    \item \textit{Level+Similarity}, which incorporates \textit{Random} and \textit{Similarity}. We add the similarity-based retrieved past learning record into the \textit{Level} prompt for behavior prediction.
\end{itemize} 

For solution simulation, we compare our beam search-based self-refinement method with two baselines:
\begin{itemize}
    \item \textit{IO}~\cite{tot}, which combines the input with task instructions and few-shot input-output examples and requires models to directly output the simulated solution;
    \item \textit{CoT}~\cite{cot}, which introduces a chain of thoughts that connects the input to the output, with each thought forming a coherent language sequence that acts as a meaningful intermediate step toward solving the problem.
\end{itemize}

Following~\citet{tot}, we demonstrate the \textit{IO} and \textit{CoT} prompt in Figure \ref{faio} and \ref{facot}.

\begin{table*}[ht]
\centering
\resizebox{\linewidth}{!}{
\begin{tabular}{c|c|ccc|ccc|ccc|ccc|ccc|ccc}
\toprule
\multicolumn{2}{c|}{\textbf{Behavior Prediction}} & \multicolumn{3}{c|}{Random} & \multicolumn{3}{c|}{Similarity} & \multicolumn{3}{c|}{Level} & \multicolumn{3}{c|}{Level+Random} & \multicolumn{3}{c|}{Level+Similarity} & \multicolumn{3}{c}{Prototype Mapping}\\
\midrule
\multicolumn{2}{c|}{\textbf{Solution Simulation}} & IO & CoT & Refine & IO & CoT & Refine & IO & CoT & Refine & IO & CoT & Refine & IO & CoT & Refine  & IO & CoT & Refine \\
\midrule\midrule

\multirow{2}[2]{*}{\textbf{\makecell[c]{LLaMA-3.3-\\70B-Instruct}}} & ROUGE-L & 12.63 & 12.23 & 18.92 & 12.74 & 12.49 & 21.32 & 12.55 & 12.39 & 20.31 & 12.49 & 13.11 & 19.28 & 13.95 & 11.88 & 20.94 & 15.76 & 13.59 & \textbf{21.37} \\
& BLEU-4 & 0.59 & 0.63 & 2.18 & 0.69 & 0.78 & 2.77 & 0.49 & 0.52 & 1.41 & 0.69 & 0.64 & 2.31 & 0.79 & 0.82 & 2.86 & 0.93 & 0.86 & \textbf{2.91}\\
\midrule

\multirow{2}[2]{*}{\textbf{\makecell[c]{Claude-3.5-\\Sonnet}}} & ROUGE-L & 21.24 & 20.96 & 23.67 & 21.8 & 25.72 & 24.61 & 18.58 & 20.99 & 22.67 & 20.24 & 20.06 & 21.51 & 23.59 & 26.55 & 26.37 & 24.5 & 24.32 & \textbf{26.63}\\
& BLEU-4 & 2.47 & 2.39 & 3.76 & 3.75 & 2.96 & 4.53 & 1.37 & 1.88 & 3.07 & 2.41 & 2.49 & 3.56 & 3.27 & 4.11 & 4.77 & 4.25 & 3.17 & \textbf{4.91}\\
\midrule

\multirow{2}[2]{*}{\textbf{\makecell[c]{GPT-3.5}}} & ROUGE-L & 23.59 & 19.94 & 25.76 & 25.59 & 19.08 & 26.34 & 20.44 & 19.23 & 24.13 & 22.68 & 19.51 & 25.41 & 22.85 & 21.38 & 26.61 & 23.88 & 20.91 & \textbf{27.37} \\
& BLEU-4 & 3.28 & 1.8 & 4.37 & 4.78 & 2.51 & 5.6 & 2.3 & 1.75 & 4.21 & 3.0 & 2.56 & 4.53 & 4.52 & 2.6 & 5.38 & 4.19 & 2.65 & \textbf{5.63}\\
\midrule

\multirow{2}[2]{*}{\textbf{\makecell[c]{GPT-4o}}} & ROUGE-L & 18.12 & 16.28 & 18.32 & 20.64 & 20.06 & 22.12 & 17.61 & 16.79 & 20.45 & 19.29 & 16.19 & 20.01 & 19.83 & 17.34 & 21.33 & 21.64 & 20.35 & \textbf{22.28} \\
& BLEU-4 & 1.77 & 1.8 & 2.44 & 2.48 & 2.37 & 3.57 & 1.29 & 1.12 & 1.8 & 2.35 & 1.54 & 2.46 & 2.8 & 2.05 & 3.26 & 3.15 & 2.55 & \textbf{4.59}\\

 \bottomrule
\end{tabular}
}
\caption{End-to-end comparison results on the \texttt{Student\_15} dataset for captioning metrics.}
\label{tmain_captioning_15}
\end{table*}

\begin{table*}
\centering
\resizebox{\linewidth}{!}{
\begin{tabular}{c|c|ccc|ccc|ccc|ccc|ccc|ccc}
\toprule
\multicolumn{2}{c|}{\textbf{Behavior Prediction}} & \multicolumn{3}{c|}{Random} & \multicolumn{3}{c|}{Similarity} & \multicolumn{3}{c|}{Level} & \multicolumn{3}{c|}{Level+Random} & \multicolumn{3}{c|}{Level+Similarity} & \multicolumn{3}{c}{Prototype Mapping}\\
\midrule
\multicolumn{2}{c|}{\textbf{Solution Simulation}} & IO & CoT & Refine & IO & CoT & Refine & IO & CoT & Refine & IO & CoT & Refine & IO & CoT & Refine  & IO & CoT & Refine \\
\midrule\midrule

\multirow{2}[2]{*}{\textbf{\makecell[c]{LLaMA-3.3-\\70B-Instruct}}} & ROUGE-L & 13.36 & 13.57 & 21.52 & 15.2 & 13.92 & 23.15 & 14.56 & 13.77 & 21.45 & 13.83 & 13.94 & 22.75 & 16.17 & 14.48 & 23.37 & 17.49 & 15.33 & \textbf{24.09} \\
& BLEU-4 & 0.6 & 0.62 & 2.08 & 0.82 & 0.81 & 2.93 & 0.65 & 0.58 & 1.45 & 0.66 & 0.64 & 2.33 & 0.92 & 0.87 & 2.99 & 0.98 & 0.93 & \textbf{3.06}\\
\midrule

\multirow{2}[2]{*}{\textbf{\makecell[c]{Claude-3.5-\\Sonnet}}} & ROUGE-L & 26.16 & 25.84 & 28.59 & 26.98 & 26.91 & 28.87 & 23.38 & 22.93 & 25.76 & 26.59 & 25.57 & 27.68 & 27.61 & 27.34 & 29.17 & 27.45 & 27.52 & \textbf{29.95} \\
& BLEU-4 & 3.29 & 2.36 & 4.98 & 4.38 & 3.31 & \textbf{5.94} & 2.35 & 1.97 & 3.31 & 3.66 & 2.83 & 4.55 & 4.81 & 3.61 & 4.94 & 4.58 & 3.18 & 5.81\\
\midrule

\multirow{2}[2]{*}{\textbf{\makecell[c]{GPT-3.5}}} & ROUGE-L & 27.45 & 22.68 & 29.15 & 28.21 & 23.49 & 30.05 & 24.39 & 22.13 & 28.95 & 26.47 & 22.63 & 28.94 & 27.14 & 24.28 & 29.86 & 27.07 & 23.44 & \textbf{30.34} \\
& BLEU-4 & 3.89 & 2.15 & 5.06 & 5.13 & 2.38 & 5.73 & 2.91 & 2.12 & 4.99 & 4.04 & 2.43 & 5.13 & 4.8 & 2.81 & \textbf{5.95} & 4.66 & 2.56 & 5.71\\
\midrule

\multirow{2}[2]{*}{\textbf{\makecell[c]{GPT-4o}}} & ROUGE-L & 22.99 & 20.96 & 22.51 & 23.97 & 22.65 & 25.15 & 18.92 & 18.67 & 21.82 & 21.14 & 19.44 & 22.2 & 22.51 & 20.77 & 23.18 & 24.9 & 22.76 & \textbf{26.6} \\
& BLEU-4 & 2.32 & 1.84 & 2.54 & 3.36 & 2.33 & 3.47 & 1.26 & 1.14 & 1.83 & 1.97 & 1.4 & 2.17 & 2.74 & 2.12 & 2.85 & 3.35 & 2.24 & \textbf{4.66} \\

 \bottomrule
\end{tabular}
}
\caption{End-to-end comparison results on the \texttt{Student\_100} dataset for captioning metrics.}
\label{tmain_captioning_100}
\end{table*}

\subsection{Metric Details}
\label{ametric}
\noindent\textbf{Accuracy}. We calculate the average accuracy of predicting whether each student correctly answers the corresponding 10 simulation records to reflect the behavior prediction quality. 

\noindent\textbf{Con\textsubscript{1}.} To assess the alignment and consistency between predicted and ground truth behavior descriptions, we introduce \textit{Con\textsubscript{1}}, a metric that utilizes a large language model (LLM) for evaluation. \textit{Con\textsubscript{1}} assigns scores from 1 to 5, with higher values indicating stronger alignment with the ground truth. For fair and reliable comparisons, all evaluations use o1-mini (API version: o1-mini-2024-09-12).

\noindent\textbf{Con\textsubscript{2}.} Similar to \textit{Con\textsubscript{1}}, \textit{Con\textsubscript{2}} employs o1-mini to assign a score to each simulated solution, with higher scores indicating better consistency with the ground truth solutions.

Detailed prompts for Con\textsubscript{1} and Con\textsubscript{2} metric are provided in Figure \ref{facon1} and \ref{facon2}.

\section{More Experimental Results and Analysis}
\label{aexp}

\subsection{End-to-end Comparison on \texttt{Student\_100}}
\label{app_student_100}
As shown in Section \ref{main_end_to_end}, we evaluate 6 behavior prediction methods and 3 solution simulation approaches, resulting in 18 unique configurations for comparison. We conduct end-to-end experiments across all configurations using the \texttt{Student\_100} dataset, which includes the whole 100 students. The results, presented in Table \ref{tmain_app}, align with the findings in Section \ref{main_end_to_end}:

1) Evaluating a student’s cognitive ability based solely on past learning accuracy is insufficient, as it does not capture mastery at the knowledge concept level. Even when selecting past records randomly or using text similarity retrieval, prediction accuracy remains low. This is because such methods often retrieve questions with similar wording but different underlying concepts, leading to misjudgments. In contrast, our cognitive prototype, constructed from a knowledge graph, accurately represents a student’s mastery of relevant concepts, enabling more precise predictions and improving solution simulation quality.

2) Given the same prototype-mapped behavior descriptions, Table \ref{tmain_app} shows that simulations based on simple IO or CoT prompts consistently underperform. This underscores the difficulty LLMs face in generating accurate solutions for students with diverse cognitive abilities in a single attempt. Our beam search-based self-refinement method addresses this challenge by iteratively improving solutions through self-evaluation and optimization. By increasing sampling frequency, it enhances the probability of generating accurate, contextually consistent solutions, leading to higher-quality simulations.

\begin{table*}[ht]
\centering
\resizebox{\linewidth}{!}{
\begin{tabular}{c|c|ccc|ccc|ccc|ccc|ccc|ccc}
\toprule
\multicolumn{2}{c|}{\textbf{Behavior Prediction}} & \multicolumn{3}{c|}{Random} & \multicolumn{3}{c|}{Similarity} & \multicolumn{3}{c|}{Level} & \multicolumn{3}{c|}{Level+Random} & \multicolumn{3}{c|}{Level+Similarity} & \multicolumn{3}{c}{Prototype Mapping}\\
\midrule
\multicolumn{2}{c|}{\textbf{Solution Simulation}} & IO & CoT & Refine & IO & CoT & Refine & IO & CoT & Refine & IO & CoT & Refine & IO & CoT & Refine  & IO & CoT & Refine \\
\midrule\midrule

\multirow{3}[2]{*}{\textbf{\makecell[c]{LLaMA-3.3-\\70B-Instruct}}}  & \textbf{Acc} & 0.46 & 0.46 & 0.46 & 0.34 & 0.34 & 0.34 & 0.34 & 0.34 & 0.34 & 0.32 & 0.32 & 0.32 & 0.34 & 0.34 & 0.34 & \textbf{0.68} & \textbf{0.68} & \textbf{0.68}\\
& \textbf{Con\textsubscript{1}} & 2.5 & 2.5 & 2.5 & 2.32 & 2.32 & 2.32 & 2.18 & 2.18 & 2.18 & 1.88 & 1.88 & 1.88 & 2.34 & 2.34 & 2.34 & \textbf{3.3} & \textbf{3.3} & \textbf{3.3}\\
& \textbf{Con\textsubscript{2}} & 2.34 & 2.4 & 1.98 & 2.2 & 2.34 & 2.04 & 2.74 & 2.14 & 1.66 & 1.86 & 2.18 & 2.0 & 2.28 & 2.48 & 2.16 & 2.78 & 2.82 & \textbf{2.84}\\
\midrule
\multirow{3}[2]{*}{\textbf{\makecell[c]{Claude-3.5-\\Sonnet}}}  & \textbf{Acc} & 0.46 & 0.46 & 0.46 & 0.54 & 0.54 & 0.54 & 0.3 & 0.3 & 0.3 & 0.46 & 0.46 & 0.46 & 0.46 & 0.46 & 0.46 & \textbf{0.56} & \textbf{0.56} & \textbf{0.56} \\
& \textbf{Con\textsubscript{1}} & 2.88 & 2.88 & 2.88 & 3.04 & 3.04 & 3.04 & 1.8 & 1.8 & 1.8 & 2.48 & 2.48 & 2.48 & 2.56 & 2.56 & 2.56 & \textbf{3.3} & \textbf{3.3} & \textbf{3.3}\\
& \textbf{Con\textsubscript{2}} & 2.46 & 2.46 & 2.8 & 2.66 & 2.8 & 2.76 & 1.86 & 1.58 & 1.32 & 2.24 & 2.0 & 2.36 & 2.5 & 2.42 & 2.4 & 2.68 & 2.44 & \textbf{2.9}\\
\midrule
\multirow{3}[2]{*}{\textbf{\makecell[c]{GPT-3.5}}}  & \textbf{Acc} & 0.46 & 0.46 & 0.46 & 0.34 & 0.34 & 0.34 & 0.46 & 0.46 & 0.46 & 0.46 & 0.46 & 0.46 & 0.42 & 0.42 & 0.42 & \textbf{0.52} & \textbf{0.52} & \textbf{0.52}\\
& \textbf{Con\textsubscript{1}} & 2.36 & 2.36 & 2.36 & 2.44 & 2.44 & 2.44 & 2.74 & 2.74 & 2.74 & 2.5 & 2.5 & 2.5 & 2.48 & 2.48 & 2.48 & \textbf{2.92} & \textbf{2.92} & \textbf{2.92}\\
& \textbf{Con\textsubscript{2}} & 2.44 & 2.78 & 3.22 & 2.56 & 2.44 & 3.02 & 2.52 & 2.84 & 2.86 & 2.54 & 2.24 & 3.18 & 2.42 & 2.44 & 2.96 & 2.68 & 2.98 & \textbf{3.28}\\
\midrule
\multirow{3}[2]{*}{\textbf{\makecell[c]{GPT-4o}}}  & \textbf{Acc} & 0.3 & 0.3 & 0.3 & 0.5 & 0.5 & 0.5 & 0.42 & 0.42 & 0.42 & 0.66 & 0.66 & 0.66 & 0.56 & 0.56 & 0.56 & \textbf{0.84} & \textbf{0.84} & \textbf{0.84}\\
& \textbf{Con\textsubscript{1}} & 2.48 & 2.48 & 2.48 & 2.9 & 2.9 & 2.9 & 2.34 & 2.34 & 2.34 & 3.56 & 3.56 & 3.56 & 3.28 & 3.28 & 3.28 & \textbf{3.8} & \textbf{3.8} & \textbf{3.8}\\
& \textbf{Con\textsubscript{2}} & 2.28 & 2.5 & 2.6 & 2.74 & 2.78 & 2.74 & 2.28 & 2.8 & 1.9 & 2.44 & 2.64 & 2.4 & 2.24 & 2.52 & 2.1 & 3.24 & 3.38 & \textbf{3.46}\\
 \bottomrule
\end{tabular}
}
\caption{End-to-end comparison on \texttt{Java\_5} dataset. \textit{Acc} and \textit{Con\textsubscript{1}} metrics both evaluate behavior descriptions, yielding identical values for the same behavior prediction method.}
\label{tjava}
\end{table*}

\begin{table*}[ht]
\centering
\resizebox{\linewidth}{!}{
\begin{tabular}{c|c|ccc|ccc|ccc|ccc|ccc|ccc}
\toprule
\multicolumn{2}{c|}{\textbf{Behavior Prediction}} & \multicolumn{3}{c|}{Random} & \multicolumn{3}{c|}{Similarity} & \multicolumn{3}{c|}{Level} & \multicolumn{3}{c|}{Level+Random} & \multicolumn{3}{c|}{Level+Similarity} & \multicolumn{3}{c}{Prototype Mapping}\\
\midrule
\multicolumn{2}{c|}{\textbf{Solution Simulation}} & IO & CoT & Refine & IO & CoT & Refine & IO & CoT & Refine & IO & CoT & Refine & IO & CoT & Refine  & IO & CoT & Refine \\
\midrule\midrule

\multirow{3}[2]{*}{\textbf{\makecell[c]{LLaMA-3.3-\\70B-Instruct}}} & \textbf{Acc} & 0.34 & 0.34 & 0.34 & 0.34 & 0.34 & 0.34 & 0.38 & 0.38 & 0.38 & 0.38 & 0.38 & 0.38 & 0.38 & 0.38 & 0.38 & \textbf{0.62} & \textbf{0.62} & \textbf{0.62}\\
& \textbf{Con\textsubscript{1}} & 2.3 & 2.3 & 2.3 & 2.16 & 2.16 & 2.16 & 1.98 & 1.98 & 1.98 & 2.2 & 2.2 & 2.2 & 2.12 & 2.12 & 2.12 & \textbf{3.0} & \textbf{3.0} & \textbf{3.0}\\
& \textbf{Con\textsubscript{2}} & 1.98 & 2.34 & 1.86 & 2.22 & 2.28 & 2.2 & 2.32 & \textbf{2.52} & 1.86 & 2.1 & 2.28 & 1.8 & 2.44 & 1.9 & 2.08 & 2.44 & \textbf{2.52} & \textbf{2.52}\\
\midrule
\multirow{3}[2]{*}{\textbf{\makecell[c]{Claude-3.5-\\Sonnet}}} & \textbf{Acc} & 0.52 & 0.52 & 0.52 & 0.62 & 0.62 & 0.62 & 0.36 & 0.36 & 0.36 & 0.46 & 0.46 & 0.46 & 0.58 & 0.58 & 0.58 & \textbf{0.66} & \textbf{0.66} & \textbf{0.66}\\
& \textbf{Con\textsubscript{1}} & 2.64 & 2.64 & 2.64 & 3.0 & 3.0 & 3.0 & 1.76 & 1.76 & 1.76 & 2.22 & 2.22 & 2.22 & 2.52 & 2.52 & 2.52 & \textbf{3.16} & \textbf{3.16} & \textbf{3.16}\\
& \textbf{Con\textsubscript{2}} & 2.62 & 2.26 & 2.26 & 2.52 & 2.36 & 2.62 & 1.94 & 1.76 & 1.48 & 1.86 & 1.88 & 2.1 & 2.46 & 2.18 & 1.88 & 2.98 & 3.1 & \textbf{3.16}\\
\midrule
\multirow{3}[2]{*}{\textbf{\makecell[c]{GPT-3.5}}} & \textbf{Acc} & 0.5 & 0.5 & 0.5 & 0.52 & 0.52 & 0.52 & 0.48 & 0.48 & 0.48 & 0.44 & 0.44 & 0.44 & 0.44 & 0.44 & 0.44 & \textbf{0.58} & \textbf{0.58} & \textbf{0.58}\\
& \textbf{Con\textsubscript{1}} & 2.62 & 2.62 & 2.62 & 2.62 & 2.62 & 2.62 & 2.36 & 2.36 & 2.36 & 2.24 & 2.24 & 2.24 & 2.3 & 2.3 & 2.3 & \textbf{2.7} & \textbf{2.7} & \textbf{2.7}\\
& \textbf{Con\textsubscript{2}} & 2.98 & 2.78 & 2.82 & 2.92 & 2.66 & 2.52 & 2.98 & 3.0 & 2.8 & 3.16 & 3.06 & 3.16 & 2.86 & 2.8 & 3.16 & 2.98 & 3.12 & \textbf{3.18}\\
\midrule
\multirow{3}[2]{*}{\textbf{\makecell[c]{GPT-4o}}}  & \textbf{Acc} & 0.5 & 0.5 & 0.5 & 0.54 & 0.54 & 0.54 & 0.42 & 0.42 & 0.42 & 0.54 & 0.54 & 0.54 & 0.68 & 0.68 & 0.68 & \textbf{0.86} & \textbf{0.86} & \textbf{0.86}\\
& \textbf{Con\textsubscript{1}} & 2.5 & 2.5 & 2.5 & 2.68 & 2.68 & 2.68 & 2.08 & 2.08 & 2.08 & 2.84 & 2.84 & 2.84 & 3.2 & 3.2 & 3.2 & \textbf{3.6} & \textbf{3.6} & \textbf{3.6}\\
& \textbf{Con\textsubscript{2}} & 2.64 & 3.04 & 3.1 & 3.14 & 3.3 & 2.9 & 2.42 & 3.08 & 2.04 & 2.32 & 2.7 & 2.24 & 2.6 & 2.8 & 2.8 & 3.32 & 3.4 & \textbf{3.42}\\
 \bottomrule
\end{tabular}
}
\caption{End-to-end comparison on \texttt{C++\_5} dataset. \textit{Acc} and \textit{Con\textsubscript{1}} metrics both evaluate behavior descriptions, yielding identical values for the same behavior prediction method.}
\label{tcplusplus}
\end{table*}

\begin{table*}[ht]
\centering
\resizebox{0.75\linewidth}{!}{
\begin{tabular}{c|ccc|ccc|ccc|ccc}
\toprule
\multirow{2}[2]{*}{\textbf{\makecell[c]{Number of\\Retrieved Tasks}}} & \multicolumn{3}{c|}{\textbf{LLaMA-3.3-70B}} & \multicolumn{3}{c|}{\textbf{Claude-3.5-Sonnet}} & \multicolumn{3}{c|}{\textbf{GPT-3.5}} & \multicolumn{3}{c}{\textbf{GPT-4o}} \\ 
\cmidrule{2-13}
 & \textbf{Acc} & \textbf{Con\textsubscript{1}} & \textbf{Con\textsubscript{2}} &
 \textbf{Acc} & \textbf{Con\textsubscript{1}} & \textbf{Con\textsubscript{2}}& 
 \textbf{Acc} & \textbf{Con\textsubscript{1}} & \textbf{Con\textsubscript{2}}& 
 \textbf{Acc} & \textbf{Con\textsubscript{1}} & \textbf{Con\textsubscript{2}} \\
 \midrule\midrule
1 & 0.61 & 2.99 & 2.69 & 0.65 & 3.09 & 2.99 & 0.56 & 2.99 & 3.49 & 0.94 & 3.77 & 3.65 \\
2 & 0.6 & 3.02 & 2.25 & 0.63 & 3.1 & 2.61 & 0.42 & 2.53 & 3.15 & 0.73 & 3.24 & 3.04 \\
3 & 0.61 & 2.97 & 2.4 & 0.62 & 3.05 & 2.72 & 0.47 & 2.63 & 3.19 & 0.7 & 3.24 & 3.12\\
 \bottomrule
\end{tabular}
}
\caption{Experiments on different number of retrieved past learning records.}
\label{tretrieve_app}
\end{table*}

\subsection{Results for Captioning Metrics}
We further evaluated the results of our solution simulation using captioning metrics, specifically ROUGE-L \cite{lin2004rouge} and BLEU-4 \cite{papineni2002bleu}. The results, shown in Tables \ref{tmain_captioning_15} and \ref{tmain_captioning_100}, indicate that our prototype mapping and self-refinement methods consistently achieve the best performance in most cases.

However, we believe that captioning metrics are not a suitable evaluation measure in the context of student simulation. The core focus of our task evaluation is to assess whether we can accurately and reasonably simulate the errors students are likely to make, rather than simply measuring text similarity. These metrics primarily evaluate low-level term similarity, which is limited in scope. Thus, they cannot effectively evaluate more complex aspects like the logical structure or syntax of code. For example, a student's incorrect code may differ from the correct version by just an indentation error, yet these metrics fail to capture such subtle distinctions. Therefore, we argue that captioning metrics do not provide a reasonable evaluation framework for our task and instead propose LLM-based metrics for assessment.

\subsection{Results on Java and C++ programming}
\label{app_java_cplusplus}
To further validate the effectiveness of our method beyond Python programming, we conduct additional experiments on two other programming subjects: Java and C++. We source new data from a new platform, CodeNet \cite{codenet}, which features tasks with a broad range of difficulty levels. Following the same expert annotation procedure of \texttt{Student\_100} described in Section \ref{text_dataset}, we construct two new datasets, \texttt{Java\_5} and \texttt{C++\_5}, each containing 5 students with 40 past learning records and 10 test records per student.

We compare our method with baselines on these 2 datasets, and the results are shown in Table \ref{tjava} and \ref{tcplusplus}. Experimental results indicate that our method consistently outperforms all baselines on both datasets. This provides strong evidence for the robustness of our framework across different programming subjects, platforms, and student profiles.

\subsection{In-Depth Analysis Details}
In section \ref{in-depth-analysis}, we analyze the impact of past learning volumes and the impact of self-refinement iterations and beam search sampling size. These experiments are conducted using the GPT-4o model and the results are shown in Figure \ref{fvolume}. We provide the detailed performance in Table \ref{tvolumes_app} and \ref{tgrid_app}.

\begin{table}[ht]
\centering
\resizebox{0.7\linewidth}{!}{
\begin{tabular}{c|ccc}
\toprule
\makecell[c]{Past Learning\\Record Volumes} & Acc & Con\textsubscript{1} & Con\textsubscript{2}\\
\midrule\midrule
10 & 0.53 & 2.82 & 2.65 \\
20 & 0.67 & 3.09 & 2.92 \\
30 & 0.77 & 3.38 & 3.1 \\
40 & 0.94 & 3.77 & 3.65 \\
\bottomrule
\end{tabular}
}
\caption{Detailed performance on different number of retrieved past learning records.}
\label{tvolumes_app}
\end{table}

\subsection{Bad Case Analysis}
\label{bad_case_app}
For student behavior prediction and solution simulation, our method is able to accurately identify the relevant knowledge concepts that a problem tests by mapping the constructed student cognitive prototype to the task in most cases. However, this pipeline may occasionally fail, particularly when careless mistakes occur in the student's solution code. These rare but unavoidable mistakes, such as typos or incorrect code indentation (which can cause compilation errors in Python), may mislead the model's predictions. For example, if such mistakes appear in the simulation records of a student with otherwise high programming abilities, our method may incorrectly predict their behavior.

In fact, these occasional careless mistakes pose a challenge not only for our method but also for all baseline approaches, as they are equally susceptible to their influence \cite{ood1, ood2}. We attribute this issue to the statistical nature of existing methods, which assess a student's cognitive level from a statistical perspective, making them ineffective at handling isolated errors.

We investigate whether increasing the number of retrieved past learning records could address this problem. The results, shown in Table \ref{tretrieve_app}, indicate that as the number of retrieved tasks increases, performance tends to decline. We speculate that retrieving more tasks introduces irrelevant information. Each retrieved task includes the problem statement, the student's solution, and expert error analysis, all of which are incorporated into the prompt for LLMs. As the number of retrieved tasks increases, the prompt becomes excessively long and redundant, overwhelming the model with too much information and leading to incorrect predictions.

Therefore, we recognize that predicting these occasional careless mistakes remains a challenging problem. Our method, however, is primarily designed to assess general cognitive proficiency—capturing a student's overall learning patterns rather than isolated, occasional errors. Fully addressing this issue would require additional methodological advancements, such as incorporating causal inference to mitigate biases and reduce the impact of confounding factors. We believe that this challenge extends beyond the scope of our current work and merits further exploration as an independent research direction.

Notably, similar challenges have been explored in other domains, such as recommendation systems~\cite{causal1, causal2, causal3}, where causal debiasing and counterfactual reasoning \cite{ood4, ood5} have been employed to address issues like clickbait. We believe that effectively handling noise in educational data is an important avenue for future research, and we plan to explore this further in subsequent work.

\subsection{Distinctions from Knowledge Tracing}
At first glance, our task appears structurally similar to Knowledge Tracing (KT) \cite{kt11, kt12}, as both involve modeling student learning processes. However, direct comparison with KT methods is not entirely appropriate due to fundamental differences in objectives and generalization capabilities:

1) Differences in objectives. Student simulation requires not only predicting whether a student answers a question correctly but also diagnosing errors in detail and simulating realistic behaviors in the form of natural language. Our method generates explicit and interpretable descriptions of mistakes and solutions, whereas KT methods focus solely on correctness prediction and lack the ability to simulate detailed behaviors or provide natural-language explanations.

2) Generalization limitations. KT models rely on implicit parametric knowledge representations and problem indices without incorporating task-specific textual inputs (\textit{e.g.}, problem statements). This reliance on training data restricts their ability to generalize to out-of-distribution (OOD) cases, such as the test data in our experiments. In contrast, our method leverages explicit cognitive prototypes, enabling robust performance in OOD scenarios and zero-shot tasks.

\begin{table}[t]
\centering
\resizebox{0.7\linewidth}{!}{
\begin{tabular}{c|ccccc}
\toprule
\diagbox[]{$B$}{$L$} & 1 & 2 & 3 & 4 & 5\\
\midrule\midrule
1 & 3.36 & 3.47 & 3.61 & 3.61 & 3.6 \\
2 & 3.46 & 3.55 & 3.65 & 3.63 & 3.61 \\
3 & 3.41 & 3.53 & 3.61 & 3.6 & 3.56 \\
\bottomrule
\end{tabular}
}
\caption{Detailed performance on different refinement iteration $L$ and beam search sampling size $B$.}
\label{tgrid_app}
\end{table}

These distinctions highlight the fundamental differences between KT and student simulation, underscoring the need for tailored methodologies when modeling student behaviors in a more interpretable and generative manner.

\stopcontents
\end{document}